\documentclass[journal,twoside]{IEEEtran}

\usepackage{cite}
\usepackage[pdftex]{graphicx}
\graphicspath{{../pdf/}{../jpeg/}}
\DeclareGraphicsExtensions{.pdf,.jpeg,.png,.eps}
\usepackage{amsmath}
\usepackage{algorithmic}
\usepackage{array}
\usepackage{stfloats}
\usepackage{url}

\usepackage{hyperref}
\usepackage{balance}
\usepackage{cleveref}
\Crefname{figure}{Fig.}{Figs.}
\Crefname{equation}{Eq.}{Eqs.}
\usepackage{siunitx}
\usepackage{mathrsfs}
\usepackage{amssymb}
\usepackage{bm}
\usepackage{subcaption}
\usepackage{threeparttable}
\usepackage{enumitem}
\usepackage[nolist,nohyperlinks]{acronym}

\usepackage[normalem]{ulem}
\usepackage{efbox,graphicx}

\renewcommand{\cref}[1]{\Cref{#1}}

\newcommand{\world}{W}
\newcommand{\base}{B}
\newcommand{\body}{\base}
\newcommand{\tool}{T}
\newcommand{\camera}{C}

\newcommand{\f}[1]{{}_{#1}}
\renewcommand{\frame}[1]{\mathscr{F}_{#1}}

\newcommand{\des}{\text{ref}}
\newcommand{\ext}{\text{ext}}
\newcommand{\act}{\text{act}}
\newcommand{\cmd}{\text{cmd}}
\newcommand{\imp}{\text{imp}}

\newcommand{\com}{\text{com}}
\newcommand{\contact}{\text{c}}
\newcommand{\scontact}{i^{\rightarrow}_{\text{c}}}
\newcommand{\econtact}{i^{\leftarrow}_{\text{c}}}

\newcommand{\spatch}{\Gamma}

\newcommand{\rot}[2]{\bm{R}_{{#1}{#2}}}
\renewcommand{\vec}[1]{\bm{#1}}
\newcommand{\torque}[1]{\bm{\tau}_{#1}}
\newcommand{\force}[1]{\bm{F}_{#1}}
\newcommand{\wrench}[1]{\widetilde{\bm{\tau}}_{#1}}
\newcommand{\wrenchest}[1]{\hat{\widetilde{\bm{\tau}}}_{#1}}
\newcommand{\mass}{m}
\newcommand{\point}{\bm{\pi}}
\newcommand{\inertia}{\bm{J}}

\newcommand{\norm}[1]{\left\lVert#1\right\rVert}
\newcommand{\eye}[1]{\mathbb{I}_{#1}}
\newcommand{\transpose}{^{\top}}
\newcommand{\real}[2]{\mathbb{R}^{{#1}\times{#2}}}

\acrodef{NDT}{\emph{non-destructive testing}}
\acrodef{UAV}{\emph{unmanned aerial vehicle}}
\acrodef{DOF}{\emph{degrees of freedom}}
\acrodef{MAV}{\emph{micro aerial vehicle}}
\acrodef{COM}{\emph{center of mass}}
\acrodef{TOF}{\emph{time of flight}}
\acrodef{ASIC}{\emph{axis-selective impedance control}}
\acrodef{VIO}{\emph{visual inertial odometry}}
\acrodef{IMU}{\emph{inertial measurement unit}}
\acrodef{PD}{\emph{proportional-derivative}}
\acrodef{PI}{\emph{proportional-integral}}

\newcommand{\ourtitle}{Active Interaction Force Control for Contact-Based Inspection with a Fully Actuated Aerial Vehicle}
\newcommand{\headertitle}{Active Interaction Force Control}

\usepackage{tikz}
\usepackage{textcomp}
\newcommand\copyrighttext{%
  \footnotesize \color{red}{\textcopyright 2020 IEEE. Personal use of this material is permitted.
  Permission from IEEE must be obtained for all other uses, in any current or future 
  media, including reprinting/republishing this material for advertising or promotional 
  purposes, creating new collective works, for resale or redistribution to servers or 
  lists, or reuse of any copyrighted component of this work in other works. 
  DOI: \href{<http://tex.stackexchange.com>}{10.1109/TRO.2020.3036623}}}
\newcommand{\cfbox}[2]{%
    \colorlet{currentcolor}{.}%
    {\color{#1}%
    \fbox{\color{currentcolor}#2}}%
}
\newcommand\copyrightnotice{%
\begin{tikzpicture}[remember picture,overlay]
\node[anchor=south,yshift=10pt] at (current page.south) {\cfbox{red}{\parbox{\dimexpr\textwidth-\fboxsep-\fboxrule\relax}{\copyrighttext}}};
\end{tikzpicture}%
}

\begin{document}
\title{\ourtitle}

\author{Karen~Bodie\IEEEauthorrefmark{1},
        Maximilian~Brunner\IEEEauthorrefmark{1},
        Michael~Pantic\IEEEauthorrefmark{1},\\
        Stefan~Walser,
        Patrick~Pf\"{a}ndler,
        Ueli~Angst,
        Roland~Siegwart,
        and~Juan~Nieto%
\thanks{\IEEEauthorrefmark{1} Authors contributed equally to this work.}%
\thanks{K. Bodie, M. Brunner, M. Pantic, S. Walser, R. Siegwart, and J. Nieto are with the Autonomous Systems Lab, ETH Z\"{u}rich, Zurich, Switzerland
\texttt{[kbodie,mabrunner,mpantic]@ethz.ch}.}%
\thanks{P. Pf\"{a}ndler and U. Angst are with the Institute for Building Materials, ETH Z\"{u}rich, Zurich, Switzerland}%
\thanks{Manuscript received December 29, 2019; revised July 24, 2020.}\vspace{-6ex}
}
\markboth{IEEE Transactions on Robotics 2020: preprint version}%
{Bodie, Brunner, Pantic, \MakeLowercase{\textit{et al.}}: \headertitle}%


\maketitle
\copyrightnotice

\begin{abstract}
This paper presents and validates active interaction force control and planning for fully actuated and omnidirectional aerial manipulation platforms, with the goal of aerial contact inspection in unstructured environments. We present a variable axis-selective impedance control which integrates direct force control for intentional interaction, using feedback from an on-board force sensor. The control approach aims to reject disturbances in free flight, while handling unintentional interaction, and actively controlling desired interaction forces. A fully actuated and omnidirectional tilt-rotor aerial system is used to show capabilities of the control and planning methods. Experiments demonstrate disturbance rejection, push-and-slide interaction, and force controlled interaction in different flight orientations. The system is validated as a tool for non-destructive testing of concrete infrastructure, and statistical results of interaction control performance are presented and discussed.

\end{abstract}


\IEEEpeerreviewmaketitle

\section{Introduction}
\label{sec:intro}

\IEEEPARstart{T}{he demand} for industrial contact inspection with aerial robots has been growing rapidly in recent years, coinciding with the development of fully actuated \acp{MAV} for aerial interaction \cite{kamel2018voliro,ollero2018aeroarms,tognon2019truly,park2018odar,ryll20196d}. A compelling and urgent case exists with aging concrete infrastructure, where a rising amount of required inspection is faced with a lack in capacity to meet the need by traditional means \cite{asce20172017}. 
Early inspection promises a more efficient and intelligent approach to long term maintenance, and a great cost savings when combined with automation.
Technologies for \ac{NDT}, such as potential mapping, permit detection of corrosion far earlier than visual assessment \cite{angst2018challenges}, but require sustained contact between the sensor and structure. 
While \acp{MAV} have been embraced as a solution for efficient visual inspection of infrastructure \cite{chan2015towards}, contact-based inspection still requires extensive human labor and the use of large supporting inspection equipment. Extending the capabilities of \acp{MAV} to perform contact inspection is the next obvious step, but also a difficult one: We now require a floating base to carry a sensor payload and to exert precise forces on the environment in any direction, while at the same time rejecting other sources of disturbance.

With new developments in inspection sensor technology, several small and light-weight devices have emerged which make \ac{MAV}-based inspection a feasible reality \cite{pfandler2019flying,isla2017emat}. 
The task remains to tackle combined interaction force control with disturbance rejection on an autonomous \ac{MAV}.
Recent research in fully actuated \acp{MAV} begins to reach this goal. The ability to exert a six \ac{DOF} force and torque allows for decoupling of the system's translational and rotational dynamics, enabling precise interaction with the environment while maintaining stability. However, making this solution a viable alternative to traditional inspection requires a system with on-board sensing, high force generation in all directions, and accurate and reliable interaction control in six \ac{DOF}.

Tilt-rotor \acp{MAV} as in \cite{allenspach2020design} can produce forces greater than gravity in many directions, offering multi- or omnidirectional flight as well as high interaction force capabilities. However, the additional complexity of such a system can increase model uncertainty.
 In addition, flying systems in general are subject to airflow disturbances (from external sources or propeller down wash), which are difficult to perceive or predict.
 Accurate control of interaction forces requires separating such disturbances and model error from interaction forces.
With recent technological improvements, force sensors have reduced in size and improved in capability, enabling direct sensing of interaction forces on a \ac{MAV} subject to various other disturbances and uncertainties. 

\subsection{Related Literature}
Interaction control techniques have been actively explored since the 1970's for fixed-base manipulators, but have not been possible for aerial robots until the past decade. Aerial interaction with traditional rotor-aligned \acp{MAV} has been achieved to varying degrees \cite{ruggiero2018aerial,hamaza20192d}, despite known limitations due to underactuation \cite{ryll20196d}.

Fully actuated \acp{MAV} are now entering the aerial robotics curriculum \cite{franchi2019interaction}, with the ability to control force and torque in six \acp{DOF} without compromising system stability. Referring to design taxonomy definitions from \cite{hamandi2020survey}, platform morphologies can be \textit{fully actuated} (having a full rank 6 \ac{DOF} allocation matrix) \cite{ryll20196d,wopereis2018multimodal}, or additionally \textit{omnidirectional} (independent of the total moment, the thrust vector can be arbitrarily directed in a spherical shell) \cite{park2018odar,bodie2018towards}. They can be fully actuated by non-parallel fixedly tilted rotors \cite{ryll20196d,park2018odar}, or with actively tilting rotor groups \cite{kamel2018voliro, ryll2014novel}.
Handling of disturbances requires their observation and has been successfully achieved on flying systems using momentum-based approaches \cite{rashad2019energy,tomic2017external}.
Several fully actuated \acp{MAV} have further performed contact inspection tasks of industrial structures \cite{trujillo2019novel,tognon2019truly}, but direct (closed-loop) force control has only been demonstrated very recently \cite{nava2019direct}.

Methods for direct force control of fixed-base manipulators are well established \cite{khatib1987unified, raibert1981hybrid}, typically switching controller modes when contact is detected. 
Switching controllers, however, are particularly unsuitable for flying systems due to the increased difficulty of contact estimation for a floating base in the presence of external disturbances.
Recent improvements for force-controlled manipulators such as intelligent collision detection \cite{haddadin2017robot} and handling of contact loss during force control \cite{schindlbeck2015unified} have seen increasing commercial adoption.
We can look to state-of-the art manipulator control techniques as inspiration for the control of newly capable fully actuated flying systems, keeping in mind the fundamental differences of a floating base system.

Spatio-temporal trajectory planning is needed to execute high-level contact inspection tasks and drive the \ac{MAV} to the surface, in contact and away again.  For flight in free space, polynomial trajectories are widely used for underactuated \acp{MAV} \cite{richter2016polynomial}. Surface-based planning for inspection and interaction has been demonstrated by extracting and connecting viewpoints based on triangulated meshes such that there is one potential view-point per mesh face  \cite{alexis2016aerial, 7140101} .

\subsection{Contributions and Extensions of Previous Work}
In this paper we present the system design and interaction control of a fully actuated aerial manipulation platform with a rigidly mounted manipulator capable of on-board computation, battery power, and sensing.

We elaborate on the contributions shown in our paper presented at RSS 2019 \cite{bodie2019omnidirectional}: The system design of a tilt-rotor \ac{MAV} with a rigid manipulator arm, a 6 \ac{DOF} \ac{ASIC} for a fully actuated flying system, and experiments showing reliable interaction control.

We present the following new contributions:
\begin{itemize}
    \item \emph{Intentional interaction control} in the form of direct force control combined with variable \ac{ASIC} for any fully actuated or omnidirectional aerial system.
    \item Statistical evaluation and comparison of results.
\end{itemize}

\section{System}
\label{sec:system}
\begin{figure}
    \begin{subfigure}{\columnwidth}
        \includegraphics[width=0.9\linewidth]{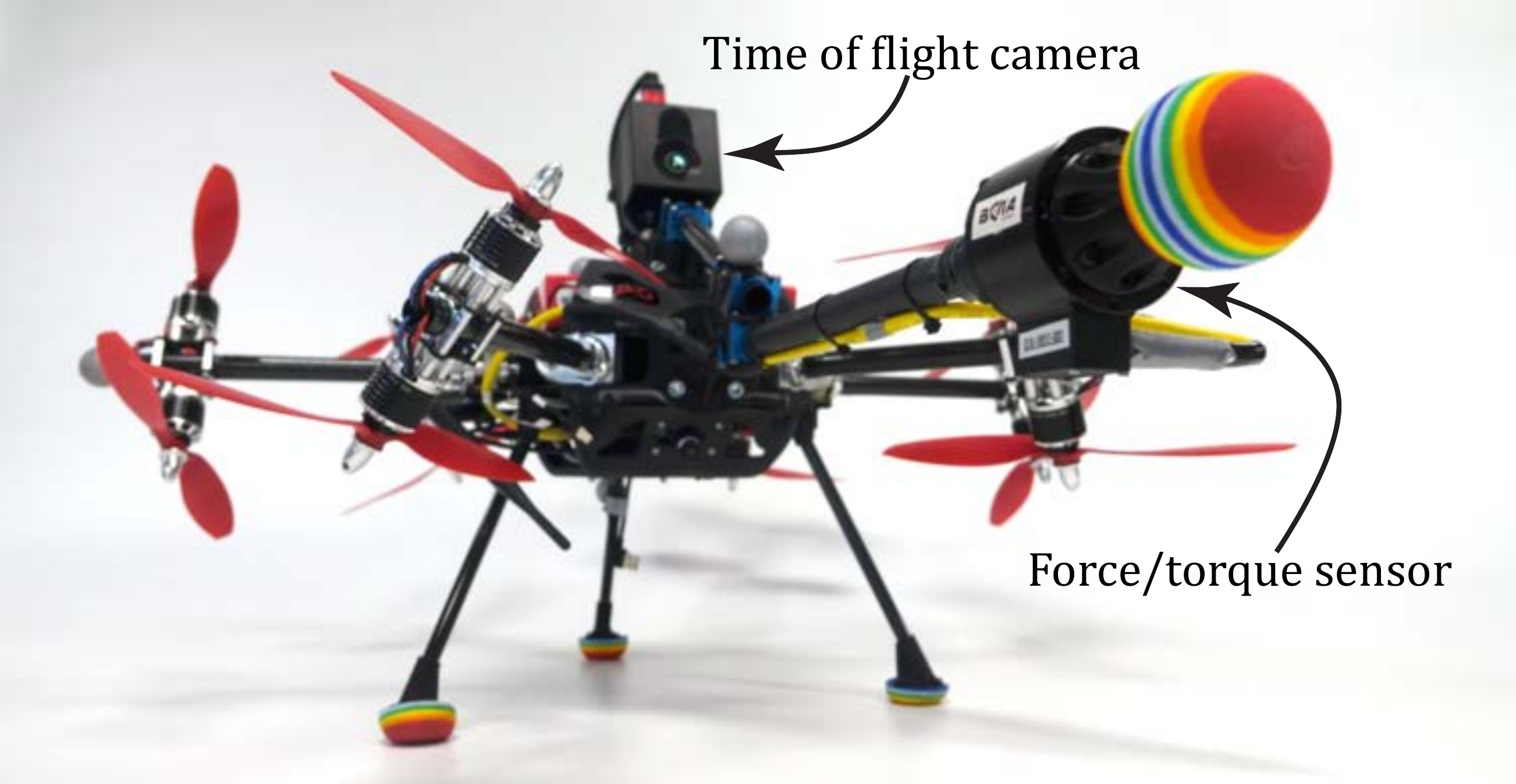}
    \centering
    \end{subfigure}
    \begin{subfigure}{\columnwidth}
       \includegraphics[width=0.8\linewidth]{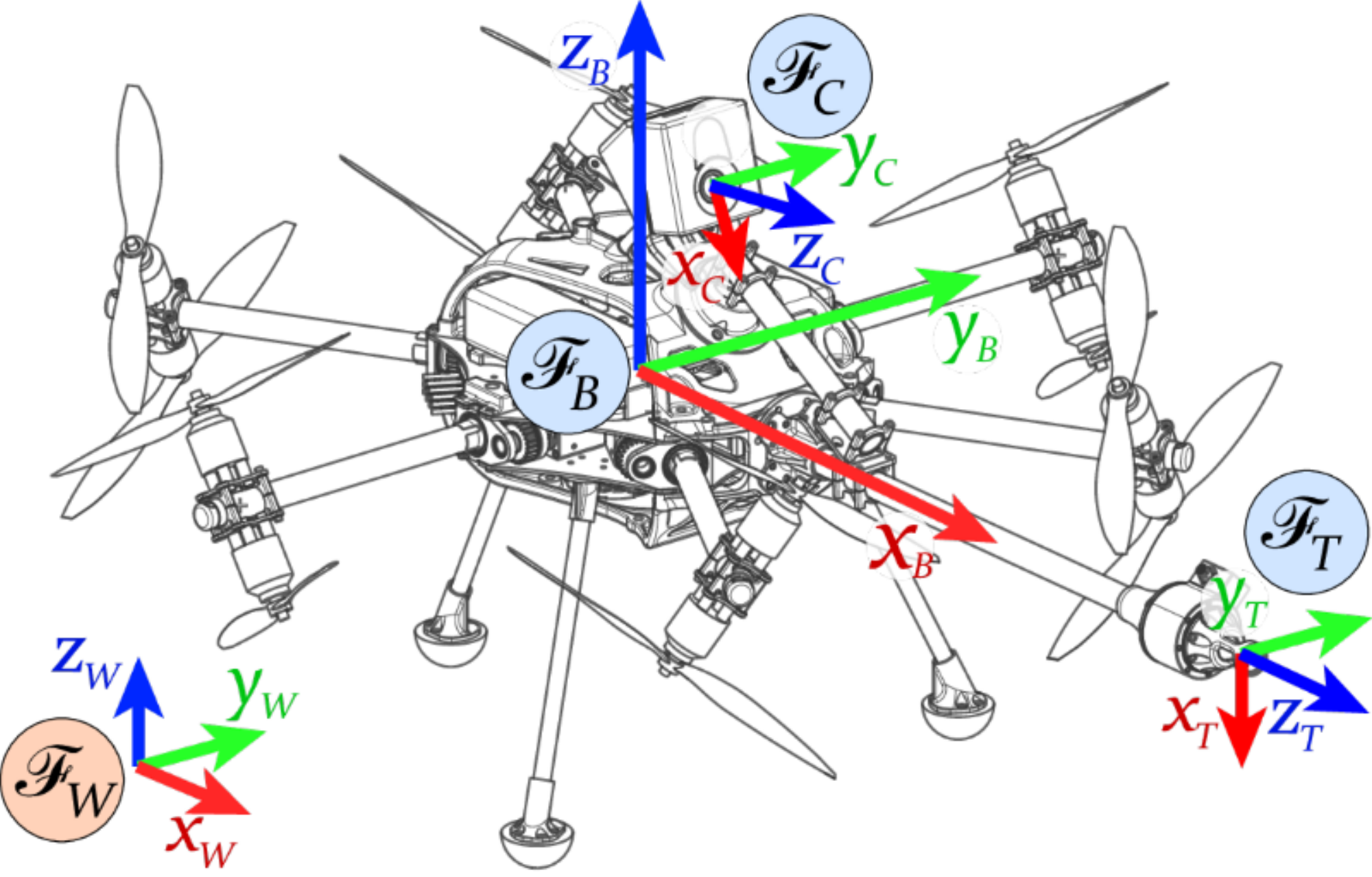}
    \centering
    \end{subfigure}
 \caption{a) System used for intentional interaction control, equipped with a 6-axis force/torque sensor near the tool tip, and b) world, base, camera and tool coordinate frames.}
 \label{fig:description}
\end{figure}

This chapter describes the \ac{MAV} system and hardware, using frame definitions presented in \cref{tab:frames} and \cref{fig:description}.

\begin{table}[htb!]
\centering
\def\arraystretch{1.2}
\begin{tabular}{l l}
\hline
\textbf{Symbol} & \textbf{Definition} \\
\hline
$\frame{*}: \{O_{*},\vec{x}_{*},\vec{y}_{*},\vec{z}_{*}\}$ & frame: origin and primary axes \\
$\world$ & inertial (world) frame subscript \\
$\base$ & body-fixed (base) frame subscript \\
$\tool$ & tool frame subscript \\
$\camera$ & TOF camera frame subscript \\
\hline
\vspace{0.1pt}
\end{tabular}
\caption{Coordinate Frame Definitions}
\label{tab:frames}
\end{table}

The \ac{MAV} used in this work takes the form of a traditional hexarotor with equally spaced arms about the $\vec{z}_\base$-axis. Each propeller group is independently tilted by a dedicated servomotor, allowing for various rotor thrust combinations.
This tilt action permits high force and torque generation in any direction, while maintaining efficient flight in horizontal hover. The resulting system is categorized as \textit{fully actuated} and \textit{omnidirectional} according to the definitions in \ref{sec:intro}.

Double rotor groups provide additional thrust for a compact size, with counter-rotating propellers to reduce gyroscopic effects. Symmetrically arranged about the tilt axis, motors also balance rotational inertia and reduce the effort required by the tilt motors. 

The platform structure is built from custom carbon fiber, aluminum, and 3D printed plastic parts. Dynamixel XL430 servomotors are used for the tilt arms, and rotors are KDE 885Kv BLDC motors with 9x4.7in propellers.
Processing occurs on an on-board Intel NUC i7 computer, and a Pixhawk low level flight controller. Two 6S \SI{3800}{\milli\ampere{}\hour} Lithium-polymer batteries are mounted for on-board power. The total system mass is \SI{5}{\kilo\gram}. Major system parameters are listed in \cref{tab:parameters}.

\begin{table}[htb!]
\vspace*{0.3cm}
\centering
\def\arraystretch{1.2}
\begin{tabular}{l c c}
\hline
\textbf{Parameter} & \textbf{Value} & \textbf{Units} \\
\hline
Total system mass & 5.0 & [\si{\kilogram}] \\
System diameter & 0.83 & [\si{\meter}] \\
Rotor group distance to $O_b$ & 0.3 & [\si{\meter}] \\ 
Maximum thrust per rotor group & 20 & [\si{\newton}] \\
Number of double rotor groups & 6 & \\
Manipulator arm length & 0.5 / 0.6 & [\si{\meter}] \\
\hline
\vspace{0.1pt}
\end{tabular}
\caption{Main system parameters}
\label{tab:parameters}
\end{table}

A manipulator arm is rigidly mounted to the platform body, with a tool frame at the tip of the arm. The $\vec{z}_\tool$-axis intersects the body origin, $O_\base$, and is collinear with the $\vec{x}_\base$-axis. Arm lengths from the body origin to the tool tip vary depending on the type of end effector, between \SI{0.5}{\meter} and \SI{0.6}{\meter}. A Picoflexx Monstar\footnote{\url{https://pmdtec.com/picofamily/monstar/}} \ac{TOF} camera is rigidly mounted near the base of the arm.

For direct force control, the system is equipped with a Rokubi 6-axis force/torque sensor\footnote{\url{https://www.botasys.com/rokubi}} on the end effector, aligned with the tool frame $\frame{\tool}$. The small size and mass (\SI{120}{\gram}) of the sensor allow integration near the tool tip, which reduces the effect of inertial and aerodynamic disturbances on the sensor measurements. With integrated EtherCAT electronics, no additional processing hardware is required. 

\section{Control Framework}
\label{sec:control}
In this section we describe a simplified system model, and introduce a novel approach for interaction control, with the goal of tracking a desired position while simultaneously generating force for interaction. The control approach extends upon a 6 DOF \ac{ASIC} as originally described in \cite{bodie2019omnidirectional}, and combines direct force and variable \ac{ASIC} which we will refer to as \emph{intentional interaction control}.

\subsection{Definitions and notation}
In the present work, we consider a general rigid-body model for a tilt-rotor aerial vehicle.
Refer to \cref{tab:symbols} for definitions of common symbols used throughout the paper.
We continue to use the frames presented in \cref{tab:frames}.

\begin{table}[htb!]
\centering
\def\arraystretch{1.2}
\begin{tabular}{l l}
\hline
\textbf{Symbol} & \textbf{Definition} \\
\hline
$\mass$ & mass\\
$\inertia$ & inertia tensor \\
$\f{A}\bm{p}_{B}$ & origin of $\frame{B}$ expressed in $\frame{A}$\\
$\rot{A}{B}$ & orientation $\in\mathrm{SO}(3)$ of $\frame{B}$ expressed in $\frame{A}$ \\
$\f{A}\vec{v}_{B}$ & linear velocity of $\frame{B}$ expressed in $\frame{A}$\\
$\f{A}\vec{\omega}_{B}$ & angular velocity of $\frame{B}$ expressed in $\frame{A}$\\
$\widetilde{\vec{v}}$ & stacked velocity vector $[\vec{v} \; \vec{\omega}]\transpose$\\
$\force{}$ & force vector\\
$\torque{}$ & torque vector\\
$\wrench{}$ & wrench vector $[\force{} \; \torque{}]\transpose \in \real{6}{1}$ \\
$\wrenchest{}$ & estimated wrench \\
$\vec{g}_W = [0 \; 0 \; g\; 0 \; 0 \; 0]^{\top}$ & gravity acceleration vector, $g = -\SI{9.81}{\meter\per\square\second}$ \\
\hline
\end{tabular}
\caption{Symbols and definitions}
\label{tab:symbols}
\end{table}

\subsection{Assumptions}
To simplify the system model, we assume that the body is rigid, and that body axes correspond with the principal axes of inertia. Thrust and drag torques are assumed proportional to squared rotor speeds, which are instantly achievable without transients. We further assume that tilt motor dynamics are negligible compared to the whole system dynamics, and tilt mechanism backlash and alignment errors are small. Airflow interference between propeller groups is assumed not to effect a significant net wrench on the system.

\subsection{System Model}
The simplified system dynamics are derived in the Lagrangian form as 
\begin{equation}
\bm{M} \dot{\widetilde{\bm{v}}} + \bm{C} \widetilde{\bm{v}} + \bm{g} = \wrench{\act} + \wrench{\ext},
\label{eq:lagrangian}
\end{equation}

\noindent where $\bm{M} \in \mathbb{R}^{6 \times 6}$ is the symmetric positive definite inertia matrix and $\bm{C} \in \mathbb{R}^{6 \times 6}$ contains the centrifugal and Coriolis terms. The terms $\wrench{\act}$ and $\wrench{\ext} \in \mathbb{R}^{6 \times 1}$ are both stacked force and torque vectors exerted on the system respectively by rotor actuation and external sources (e.g. contact or wind disturbances). 
When expressed in the body fixed frame $\frame{\base}$, 

\newcommand\numberthis{\addtocounter{equation}{1}\tag{\theequation}}
\begin{align*}
\bm{M} & = \text{diag} \left( \begin{bmatrix} m \eye{3} & \bm{J}\end{bmatrix} \right) \\
\bm{C} & = \text{diag} \left( \begin{bmatrix} m [_{\body}\bm{\omega}_{\body}]_{\times} & -\bm{J} [_{\body}\bm{\omega}_{\body}]_{\times} \end{bmatrix} \right) \numberthis\\
\bm{g} & = m \: \text{diag} \left( \begin{bmatrix} \rot{B}{W} & \bm{0}_{3\times 3}\end{bmatrix} \right) \bm{g}_W,
\end{align*}

\noindent where $[\bm{*}]_{\times}$ is the skew-symmetric matrix associated with vector $\bm{*}$.

\begin{align}
\begin{split}
    \vec{e}_{p} &= \rot{\base}{\world}(\f{\world}\vec{p}-\f{\world}\vec{p}_{\des})\\
    \vec{e}_{R} &= \frac{1}{2}\left( \rot{\world}{\base,\des}\transpose\rot{\world}{\base}-\rot{\world}{\base}\transpose\rot{\world}{\base,\des}\right)^\vee \\
    \vec{e}_{v} &= \f{\base}\vec{v}-\rot{\base}{\world}\f{\world}\vec{v}_{\des}\\
    \vec{e}_{\omega} &= \f{\base}\vec{\omega}_{\world\base}-\rot{\base}{\world}\,\f{\world}\vec{\omega}_{\world\base,\des},
\label{eq:error}
\end{split}
\end{align}
where we use the \emph{vee}-operator $*^\vee$ to extract a vector from a skew symmetric matrix. We then write the stacked error vectors as
\begin{align}
\begin{split}
    \widetilde{\vec{e}}_{p} &= [\vec{e}_{p}\transpose\; \vec{e}_{R}\transpose]\transpose \in \real{6}{1}\\
    \widetilde{\vec{e}}_{v} &= [\vec{e}_{v}\transpose\; \vec{e}_{\omega}\transpose]\transpose \in \real{6}{1}.
\end{split}
\end{align}

Tracking error terms are defined in $\frame{\base}$ in \eqref{eq:error}. The trajectory is transformed from the inertial frame to compute the error, and we stack the resulting pose and generalized velocity errors.

\subsection{External Wrench Estimation}
In order to account for the influence of contact forces, we employ an external wrench estimator using a generalized momentum approach. Our implementation follows the method described in \cite{ruggiero2014impedance}, and is expressed as

\begin{equation}
\wrenchest{\ext} = \bm{K}_I\left(\bm{M}\widetilde{\bm{v}}- \int\left(\wrench{\cmd} - \bm{C}\widetilde{\bm{v}} - \bm{g} + \wrenchest{\ext}\right)\text{dt}\right),
\label{eq:wrench_estimation}
\end{equation}

\noindent where we assume that the commanded $\wrench{\cmd}$ is able to achieve the desired actuation wrench $\wrench{\act}$. The positive definite diagonal observer matrix $\bm{K}_I \in \real{6}{6}$ acts as an estimator gain. Differentiating \eqref{eq:wrench_estimation}, a first-order low-pass filtered estimate $\wrenchest{\ext}$ of the external wrench $\wrench{\ext}$ is obtained:

\begin{equation}
\dot{\hat{\widetilde{\bm{\tau}}}}_{\ext} = \bm{K}_I(\wrench{\ext} - \wrenchest{\ext}).
\label{eq:firstOrderLowPass}
\end{equation}

Note that \eqref{eq:wrench_estimation} allows estimation of external forces and torques without the use of acceleration measurements, only requiring linear and angular velocity estimates.

\begin{figure*}[tp]
\centering
\includegraphics[trim=2.5cm 6cm 1cm 19cm, clip, width=0.85\textwidth]{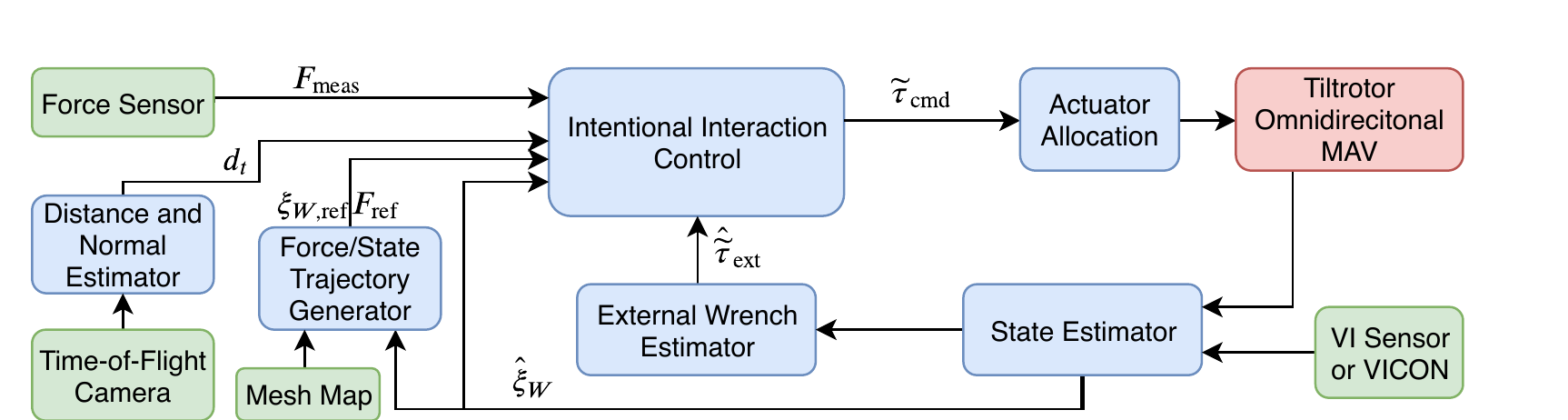}
 \caption{Block diagram of the intentional interaction control framework, combining direct force control with variable \ac{ASIC}. The full pose estimate and pose reference are represented by $\hat\xi_W$ and $\xi_{W,ref}$, respectively.}
 \label{fig:force_control_diagram}
\end{figure*}

\subsection{Intentional Interaction Control}
\label{sec:intentional_interaction}
In this section, we propose a novel control approach which we refer to as \textit{intentional interaction control}. This extends upon our previous work of \ac{ASIC} with two additional considerations.

The first extension uses distance sensing for variable compliance as a function of the end effector’s perceived distance from a surface. This context-based approach addresses the problem of amplified model error and disturbances along the end effector axis, even when no interaction surface is present.

The second extension incorporates direct force control when a desired force is prescribed. The momentum-based wrench estimate used for impedance control contains an accumulation of force and torque unrelated to the point of interaction, and in many cases cannot be used for direct force tracking. The concept presented here takes advantage of a multi-axis force sensor mounted at the tool tip to resolve differences in interaction forces and other aerial disturbances. This approach offers an improvement upon impedance control in its ability to track a desired force with direct force sensor feedback.

The resulting control approach embodies the higher level idea of purposeful interaction, and should be used with a planner that is aware of its environment and the interaction task. Force control is only attempted when it is explicitly communicated by the planner, and motion tracking is performed at all times. At points of interaction, the desired tool trajectory should trace the surface, and an additional vector provides a desired force command. When free flight is intended, this interaction force component is zero. A certain extent of planner error is handled by an interaction \textit{confidence factor}, as described below.

\subsubsection{Variable Axis-Selective Impedance Control}\label{sec:selective_impedance}
As previously derived in \cite{bodie2019omnidirectional}, the system's full actuation enables us to implement an impedance controller with virtual inertia that can be individually selected on each axis in 6 \ac{DOF}. Selecting the values to be higher or lower than the real inertial parameters causes the system to reject disturbances in some directions while exhibiting compliant behavior in others.
We choose an impedance control scheme with the following desired closed loop system dynamics:
\begin{equation}
\bm{M}_v\dot{\widetilde{\bm{v}}} + \bm{D}_v\widetilde{\vec{e}}_{v} + \bm{K}_v\widetilde{\vec{e}}_{p} = \wrench{\ext},
\label{eq:desired_dynamics}
\end{equation}

where $\bm{M}_v, \bm{D}_v,$ and $\bm{K}_v \in \mathbb{R}^{6 \times 6}$ are positive definite matrices representing the desired virtual inertia, desired damping, and desired stiffness of the system. Here we consider the desired case where $\wrenchest{\ext} = \wrench{\ext}$.

\begin{figure}[htp]
\centering
\includegraphics[width=\columnwidth]{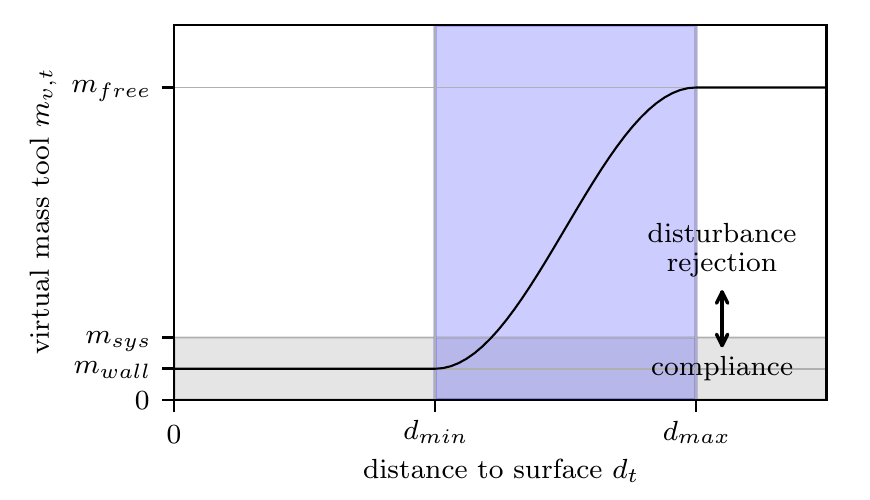}
 \caption{Variable impedance in the $\vec{z}_{\tool}$ direction as a function of wall distance $d_t$, with transition zone shown in blue. Virtual mass is shown in relation to the total system mass $m_{sys}$.}
 \label{fig:dist_vs_vmass}
\end{figure}

We address the issue of constant compliance to disturbances and error in the end effector direction by choosing $\bm{M}_v$ as a function of the distance measurement $d_t$ of a surface from the tool frame along $\vec{z}_{\tool}$, as described in \cref{sec:distanceest}. In free flight, when surfaces are far away, the virtual mass $\mass_{v, \text{free}}$ along the end effector axis is as high as all other axes to reject disturbances. Within a range from $d_{\text{max}}$ to $d_{\text{min}}$ from the surface, that value ramps down to $\mass_{v, \text{wall}}$  according to a sine function to exhibit compliance (\cref{fig:dist_vs_vmass}). The resulting virtual mass is calculated as follows:

\begin{align}
c_{v,\tool}(d_t) &=
    \begin{cases}
    1, & \text{if } d_t \leq d_{\text{min}}\\
    0.5 (1 + \cos(\frac{d_t - d_{\text{min}}}{d_{\text{max}} - d_{\text{min}}} \pi)), & \text{if } d_{\text{min}} < d_t \leq d_{\text{max}}\\
    0,               & \text{otherwise}.
    \end{cases} \\
\mass_{v,\tool}(d_t) &= c_{v,\tool}(d_t)(\mass_{v,\text{free}} - \mass_{v,\text{wall}}) + \mass_{v,\text{wall}}\nonumber
\\
\begin{split}
    \bm{M}^*_v(d_t) &= \\ 
    &\text{diag}\left(
    \begin{bmatrix}
    \mass_{v,\text{free}}&
    \mass_{v,\text{free}}&
    \mass_{v,\tool}(d_t) &
    J_{v} & J_{v} & J_{v}\end{bmatrix}
    \right)
    \end{split}
   \nonumber
\\
    \bm{M}_v(d_t) &= \bm{R} \ \bm{M}^*_v(d_t) \ \bm{R}\transpose,
     \label{eq:virtual_mass}
\end{align}

\noindent where $\bm{R} = \text{blockdiag}\left(\bm{R}_{\base\tool}, \bm{R}_{\base\tool}\right)$.
Note that the virtual inertia $J_v$ is equal in all axes in order to reject torque disturbances and to track the desired attitude while in contact.

We can then derive the applied control wrench by substituting $\dot{\widetilde{\bm{v}}}$ from \eqref{eq:desired_dynamics} into \eqref{eq:lagrangian} as follows:

\begin{equation}
    \begin{split}
    \wrench{\cmd} = &
    (\bm{M}\bm{M}_v(d_t)^{-1} - \mathbb{I}_6)\wrenchest{\ext} \\
    & - \bm{M}\bm{M}_v(d_t)^{-1}(\bm{D}_v\widetilde{\vec{e}}_{v} + \bm{K}_v\widetilde{\vec{e}}_{p}) + \bm{C}\widetilde{\bm{v}} 
    + \bm{g}.
    \end{split}
\label{eq:impedance1}
\end{equation}

For readability, we hereafter refer to $\bm{M}_v$ and related variables without explicitly stating their dependency on $d_t$.

Defining normalized inertia, stiffness and damping values respectively as $\overline{\bm{M}}_v = \bm{M}^{-1}\bm{M}_v$, $\overline{\bm{D}}_v = \overline{\bm{M}}_v^{-1}\bm{D}_v$ and $\overline{\bm{K}}_v = \overline{\bm{M}}_v^{-1}\bm{K}_v$, effective stiffness and damping of the real system are represented by $\overline{\bm{D}}_v$ and $\overline{\bm{K}}_v$. These values are experimentally tuned for $\overline{\bm{M}}_v = \eye{6}$ (no influence of the external wrench estimate) and set as constant positive-definite matrices. The desired system dynamics $\bm{D}_v$ and $\bm{K}_v$ then change as a function of $\bm{M}_v$.

\begin{equation}
\begin{matrix}
\bm{D}_v(\bm{M}_v) = \overline{\bm{M}}_v\overline{\bm{D}}_v = \bm{M}^{-1}\bm{M}_v \overline{\bm{D}}_v \\
\bm{K}_v(\bm{M}_v) = \overline{\bm{M}}_v\overline{\bm{K}}_v = \bm{M}^{-1}\bm{M}_v \overline{\bm{K}}_v.
\end{matrix}
\end{equation}

We then rewrite \eqref{eq:impedance1} as
\begin{equation}
    \wrench{\cmd} = (\overline{\bm{M}}_v^{-1} - \eye{6})\wrenchest{\ext} - \overline{\bm{D}}_v\widetilde{\vec{e}}_{v} - \overline{\bm{K}}_v\widetilde{\vec{e}}_{p} + \bm{C}\widetilde{\bm{v}} + \bm{g}.
    \label{eq:impedance2}
\end{equation}

\subsubsection{Direct force control}
We aim to apply direct force control in order to exert a specific force onto a surface. The reference force is given by a trajectory $\force{\des}(t)$ that is defined according to the task.
A force trajectory is only acted upon when there is confidence that the reference force can exist at a location near enough to the target. This confidence is a function of the perceived surface distance $d_t$ and the tool position error $\bm{e}_t$ in the direction of desired force \eqref{eq:projected_force}, as calculated in \eqref{eq:lambda}. The value is smoothly transitioned with a first-order filter with coefficient $c_{\lambda}$ to avoid step inputs at the start or end of a non-zero desired force. We arrive at the computation of the confidence factor $\lambda_k$ which represents $\lambda$ at time step $k$.

\begin{equation}
    \bm{e}_t = \bm{p}_t - \bm{p}_{t,\text{\des}} \in \real{3}{1}
\end{equation}
\begin{equation}
    e_{t,\force{\des}} = \frac{\bm{e}_t\cdot\force{\des}}{\norm{\force{\des}}}
    \label{eq:projected_force}
\end{equation}

\begin{align}
    \lambda_{d} &= 
    \begin{cases}
    1, & \text{if } d_t \leq d_{\text{min}}\\
    0.5 (1 + \cos(\frac{d_t - d_{\text{min}}}{d_{\text{max}} - d_{\text{min}}} \pi)), & \text{if } d_{\text{min}} < d_t \leq d_{\text{max}}\\
    0,               & \text{otherwise}.
    \end{cases}
    \\
    \lambda_{e} &=
    \begin{cases}
    1, & \text{if } e_{t,\force{\des}} \leq e_{\text{min}}\\
    0.5 (1 + \cos(\frac{e_{t,\force{\des}} - e_{\text{min}}}{e_{\text{max}} - e_{\text{min}}} \pi)), & \text{if } e_{\text{min}} < e_{t,\force{\des}} \leq e_{\text{max}}\\
    0,               & \text{otherwise}.
    \end{cases}
    \\
    \lambda_k &= 
    \begin{cases}
    c_{\lambda} \lambda_{d} \lambda_{e} + (1 - c_{\lambda})\lambda_{k-1}, & \text{if } \norm{\force{\des}} > 0 \\
    0,               & \text{otherwise}.
    \end{cases}
\label{eq:lambda}
\end{align}

\begin{figure}[tp]
\centering
\includegraphics[trim=0cm 0cm 0cm 0cm, clip, width=\columnwidth]{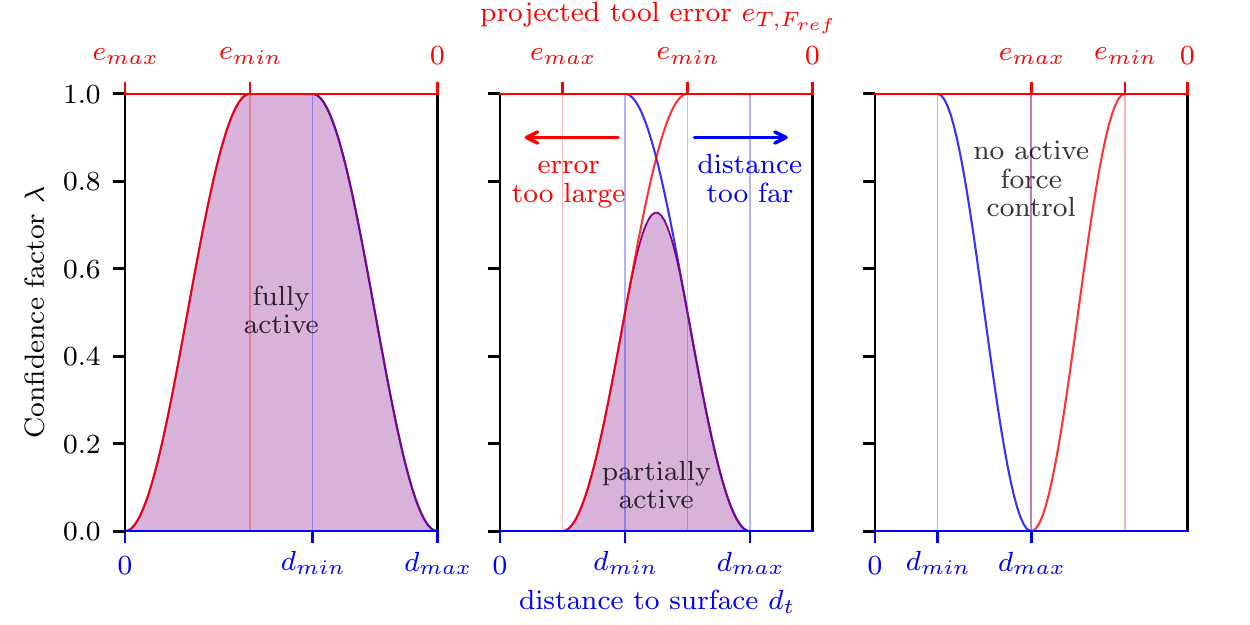}
 \caption{Confidence factor $\lambda$ as a function of wall distance $d_{\tool}$, and projected tool error $e_{t,\force{\des}}$. The solid area indicates the proportion of force control used.}
 \label{fig:lambda}
\end{figure}

The behavior of the combined force and impedance control is shown in \cref{fig:lambda}. In the nominal case, the planner commands a path in free flight which the controller is able to achieve, and a desired force is commanded only when the sensed distance $d_t$ and projected tool error $e_{t,\force{\des}}$ are small. In the case where the set point is behind the wall, the controller uses compliant impedance control in the direction of the end effector to perform its task as well as possible. When a force is then commanded,  $\lambda$ is $1$ and force control is fully active. When the set point is in front of the wall between $d_\text{min}$ and $d_\text{max}$, there is a transition phase where $\lambda$ is between 1 and 0, a compromise between trying to achieve force control and maintain trajectory tracking. In the case where the wall is not sensed within $d_\text{max}$, $\lambda$ is $0$ and no force control is attempted.

\subsubsection{Unified wrench command}

\begin{figure}[tp]
\centering
\includegraphics[trim=4.5cm 18cm 4.5cm 7cm, clip, width=0.9\columnwidth]{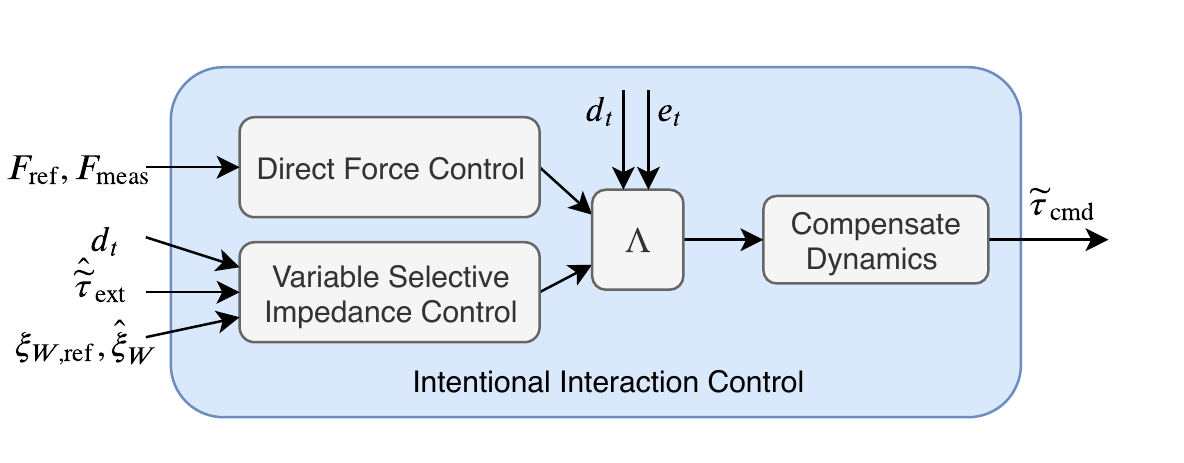}
 \caption{Diagram of intentional interaction control, with impedance control varying the virtual mass along the tool axis as a function of the sensed distance $d_t$. Direct force control is added selectively as a function of $d_t$ and the tool error $\bm{e}_t$.}
 \label{fig:force_control_detail}
\end{figure}

We define a selection matrix $\bm{\Lambda}$, which orients the confidence factor $\lambda$ from \eqref{eq:lambda} in the direction of desired force $\force{\des}$ using the rotation $\rot{\force{\des}}{z}$, and which is constructed as follows:

\begin{align}
    \bm{\Lambda} &= \rot{\force{\des}}{z}
        \begin{bmatrix}
        0 & 0 & 0 \\ 
        0 & 0 & 0 \\ 
        0 & 0 & \lambda
        \end{bmatrix},\;
        \rot{\force{\des}}{z}\transpose \in \real{3}{3}
    \label{eq:force_selection_matrix} \\
    \widetilde{\bm{\Lambda}} &= \text{blockdiag}\left(\bm{\Lambda}, \bm{0}_{3,3}\right) \in \real{6}{6}.
\end{align}

We augment this matrix to 6 DOF with zeros since we have a point end effector, but this could be extended for intentional interaction torques. The matrix positively selects direct force control commands. We use a \ac{PI} control scheme with a feed forward term to track the given reference force $\f{\base}\force{\des}$ based on the force tracking error $\bm{e}_f$. The interaction force $\force{t}$ at the end effector is measured by an on-board sensor.

\begin{align}
   \bm{e}_f &= \force{t}-\force{\des}\\
   \force{\text{dir}} &= \frac{1}{\mass}\bm{\Lambda}\left(-\force{\des,\base} + \bm{K}_{f,p} \bm{e}_f
   + \bm{K}_{f,i} \int\bm{e}_f \text{dt}\right) \\
   \wrench{\text{dir}} &= \begin{bmatrix} \force{\text{dir}}^T & 0 & 0 & 0\end{bmatrix}^T \in \mathbb{R}^{6\times 1}.
\end{align}

Using $\bm{M}_v$ from \eqref{eq:virtual_mass} and normalizing by the system mass as before, we compute the impedance control command as

\begin{equation}
    \wrench{\text{imp}} = (\eye{6} - \widetilde{\bm{\Lambda}})(\bm{R}^\top \overline{\bm{M}}_v^{-1} \bm{R} - \eye{6})\hat{\wrench{}}_{\text{ext}}  - \overline{\bm{D}}_v\widetilde{\vec{e}}_{v} - \overline{\bm{K}}_v\widetilde{\vec{e}}_{p},
    \label{eq:impedance_command}
\end{equation}

\noindent where the selection matrix counterpart $(\eye{6} - \widetilde{\bm{\Lambda}})$ is used to remove the component of the momentum-based wrench estimate in the direction of desired interaction.

The two control commands are then combined and compensated for nonlinear dynamic effects and gravity. Since the system \ac{COM} is not located at the geometric center of control, we use a feed forward term to compensate for the torque $\torque{\com}$ caused by the offset $\f{\base}\vec{p}_{\com}$. The final resulting wrench command is shown in \eqref{eq:combined_command}, and shown as a block diagram in \cref{fig:force_control_detail}.

\begin{align}
    \wrench{\cmd}^* &= \wrench{\text{dir}} + \wrench{\imp} + \bm{C}\widetilde{\bm{v}} + \bm{g}
    \\
    \wrench{\com} &= 
    \begin{bmatrix} 
        \bm{0}_{3\times1} \\
        \f{\base}\vec{p}_{\com}\times\f{\base}\force{\cmd}
    \end{bmatrix}
    \\
    \wrench{\cmd} &= \wrench{\cmd}^* + \wrench{\com}.
    \label{eq:combined_command}
\end{align}

\subsection{Vehicle center of mass estimation}
Inertial tensor values and an initial \ac{COM} estimate for the system come from a detailed CAD model, which we verify as having the same total mass as our measured system. These inertia values (an average of the extreme rotor alignment configurations) are used directly in the controller.

The center of mass is subject to change depending on the platform configuration that is used for a specific application. Since our initial estimate of the \ac{COM} based on the system architecture directly (e.g. a CAD model) is not exact enough to eliminate steady state pose error, we use a calibration procedure to estimate the \ac{COM}. During this procedure, the platform performs a trajectory consisting of pitching and rolling while hovering in a constant position.
Assuming hover in steady state (i.e. negligible angular accelerations and velocities), we can write the balance of a \ac{COM} offset and the commanded torque as
\begin{equation}
    \f{\base}\vec{p}_{\com}\times \f{\base}\force{\cmd} = \f{\base}\torque{\cmd}.
\end{equation}
During a calibration flight we record a dataset of $N$ datapoints of commanded forces $\bar{\vec{F}}=\{\f{\base}\force{1},\dots,\f{\base}\force{N}\}$ and torques $\bar{\vec{\tau}}=\{\f{\base}\torque{1},\dots,\f{\base}\torque{N}\}$. We then use linear least squares optimization to solve for $\f{\base}\bm{p}_{\com}$:
\begin{equation}
    \f{\base}\vec{p}_{\com} = (\vec{X}^\top\vec{X})^{-1}\vec{X}^\top\vec{y},
\end{equation}
with
\begin{equation}
    \vec{X}=\begin{bmatrix}
    -[\f{\base}\force{1}]_\times\\
    \vdots\\
    -[\f{\base}\force{N}]_\times
    \end{bmatrix},\quad\vec{y}=\begin{bmatrix}
    \f{\base}\torque{1}\\
    \vdots\\
    \f{\base}\torque{N}
    \end{bmatrix}.
\end{equation}

We acknowledge that this \ac{COM} estimate includes additional effects due to model uncertainties, but performance improvement following calibration suggests that our \ac{COM} compensation model captures the majority of these steady-state effects.

\subsection{Force Sensor Filtering}
Any additional mass on the end of the force sensor will lead to inertial forces and torques from dynamic movement. This can be modeled with known pose of the end effector and static and dynamic parameter identification, but in our case is ignored due to very low mass. We still expect to see noise from vibration, which can be addressed with a filter.

We use a 2nd-order low pass butterworth filter with a cutoff frequency of \SI{5}{\hertz}. This filter yields smooth force measurements with reasonable latency for surface inspection tasks.

\section{Distance Estimation and Planning for Interaction}
\label{sec:planning}
\subsection{Surface Distance Estimation}
\label{sec:distanceest}

For surface distance estimation the predicted contact point is defined as the intersection of an observed surface and the $\vec{z}_\tool$-axis.
In order to estimate the distance to this point, a dense point cloud $\f{C}\Pi = \{\f{C}\point^{0} \dots \f{C}\point^{N}\}$ is obtained from the \ac{TOF} camera and all points within a certain distance $\f{T}d_{\pi}$ to the $\vec{z}_\tool$-axis are selected. This subset $\f{T}\bm{\spatch}$ is formalized as:
\begin{equation}
       \f{T}\bm{\spatch} = \{\f{T}\point^{i}\ |\ \f{C}\point^{i} \in \f{C}\Pi\ \land
\left \lVert \f{T}\point^{i} \times (\f{T}\point^{i} - \mathbf{e}_{z}) \right \lVert \leq \f{T}d_{\pi}  \},
\end{equation}
where $\f{T}\point^{i}$ is a point of the point cloud transformed into the tool frame $\frame{\tool}$ according to
\begin{equation}
   \f{T}\point^{i} = \rot{T}{C} \cdot \f{C}\point^{i} + \f{T}\bm{p}_{C},
\end{equation}
and $\vec{e}_{z} = \left[0\ 0\ 1 \right]^{T}$ is the unit vector of the $z$ axis.

The surface patch center $\f{T}\bm{\spatch}_{\otimes}$ is then obtained by the unweighted average of all points belonging to $\f{T}\bm{\spatch}$:
\begin{align}
    M &= \lvert \f{T}\bm{\spatch} \rvert \\
    \f{T}\bm{\spatch}_{\otimes} &= \frac{1}{M}\sum_{i=0}^{M} \f{T}\point^{i},\ \f{T}\point^{i} \in  \f{T}\bm{\spatch}.
\end{align}

The surface patch normal $\f{T}\bm{\spatch}_{\perp}$ is obtained by plane fitting via singular value decomposition on the un-biased points $\f{T}\widetilde{\point}^{i}$.
\begin{align}
    \f{T}\widetilde{\point}^{i} &= \f{T}\point^{i} -  \f{T}\bm{\spatch}_{\otimes}\\
    \bm{U}\bm{\Sigma}\bm{V} &= \text{SVD}\left(\begin{bmatrix}
    \f{T}\widetilde{\point}^{0}, \dots, \f{T}\widetilde{\point}^{M})
    \end{bmatrix}\transpose\right)
\end{align}
The resulting $\bm{V}$ is a $3 \times 3$ matrix, and the un-normalized surface patch normal corresponds to the last column:
\begin{align}
    \bm{V} &= \begin{bmatrix}\cdot, \cdot, \f{T}\bm{\spatch'}_{\perp} \end{bmatrix} \\
    \f{T}\bm{\spatch}_{\perp} &= \f{T}\bm{\spatch'}_{\perp} \cdot  \frac{1}{ \lVert\f{T}\bm{\spatch'}_{\perp} \rVert}
\end{align}

Furthermore, the surface patch normal sign is corrected such that
$\vec{e}_{z} \cdot  \f{T}\bm{\spatch}_{\perp}$ is always negative.

\begin{figure}[h]
\centering
  \includegraphics[clip,width=0.48\textwidth]{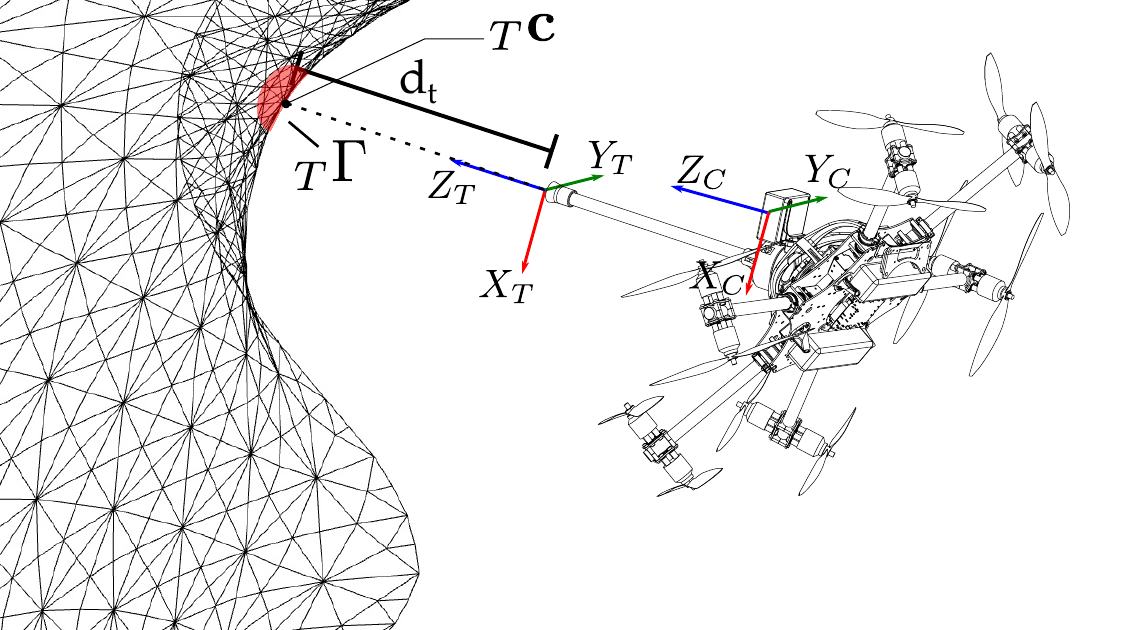}
\caption{Distance and normal estimation: $\f{\tool}\vec{c}$ is the contact point defined by the intersection of the $\vec{z}_{\tool}$-axis and the observed surface. $d_{t}$ depicts the distance to the contact point, and $\f{T}\bm{\spatch}$ is the set of 3D points used to estimate normal and distance.}
\label{fig:planning}
\end{figure}

Finally, $\vec{c}_\tool$ is obtained by intersecting the tool axis $\vec{e}_{z}$ and the obtained surface patch plane $\{\f{T}\bm{\spatch}_{\otimes}, \f{T}\bm{\spatch}_{\perp}\}$ as follows:
\begin{align}
    \gamma &= \frac{\f{\tool}\bm{\spatch}_{\otimes}\ \cdot\ \f{\tool}\bm{\spatch}_{\perp}}{\vec{e}_{z}\ \cdot \ \f{T}\bm{\spatch}_{\perp}} \\
    \f{\tool}\vec{c} &= \gamma \cdot \vec{e}_{z}.
\end{align}

For the experiments in this work $\f{T}d_{\pi}$ is chosen to be $\SI{0.1}{\meter}$, which yields approximately $2600$ points for distance and normal estimation when in contact and about $500$ at a distance of $d_{t}=\SI{1}{\meter}$. By averaging this large amount of individual measurements per estimate, we obtain a smooth, exact and very low noise distance measurement.

The resulting distance measurement is fed to the controller in real-time, whereas the following trajectory planning is executed in advance with a reconstructed surface map, if available.
\subsection{Force Trajectory Planning}
\label{sec:meshplanning}
Interacting with a surface requires knowledge about the location and orientation of the surface. 
We use the \ac{TOF} camera to obtain a point cloud of the surface to interact with, which is then converted to a triangular mesh with face and vertex normals by using Poisson surface reconstruction \cite{kazhdan2006poisson}. In order to obtain a contact point that lies exactly on the reconstructed mesh and that is closes to a desired free-space point, we use efficient AABB-tree lookups\cite{cgal:atw-aabb-20a}. The tool is then driven to the resulting contact point $\f{W}\bm{p}_{\tool,\contact}$ and aligned with its surface normal $\f{W}\bm{p}_{\perp,\contact}$ by planning a body trajectory in world frame,
\begin{multline*}
\f{W}\Theta\f{B} = \{\f{W}\bm{p}\f{B}^{i}, \rot{W}{B}^{i}, \f{W}\vec{v}_{B}^{i}, \f{W}\vec{\omega}_{B}^{i}, \f{W}\force{B}^{i}, \f{W}\torque{B}^{i}  \},\\ i=0\dots100t,
\end{multline*}
where $t$ is the duration of the trajectory in seconds. In the following, $\scontact$ is the index of the first, and $\econtact$ of the last set-point that is in contact with the surface.

We use polynomial trajectory interpolation \cite{richter2016polynomial} and non-linear optimization as described in \cite{burri2015real-time} to obtain smooth trajectories for $\f{W}\bm{p}\f{B}^{i}, \rot{W}{B}^{i}, \f{W}\vec{v}_{B}^{i} \text{ and } \f{W}\vec{\omega}_{B}^{i}$ between specific constrained set-points, such as start, contact point and end. The polynomial trajectory is then sampled at $\SI{100}{\hertz}$ and passed to the controller as an array of timestamped set-points.
In the following, the construction of the contact point constraints $\f{W}\Theta\f{B}^{i}\ \text{for}\ i=\scontact \dots \econtact$ are described in detail. As a shorthand for static properties during contact, the index $^{\contact}$ is used. The body position $\f{W}\bm{p}\f{B}^{\contact}$ that causes the tool to contact the desired location is obtained by standard frame transformations.
\begin{equation}
    \f{W}\bm{p}\f{B}^{\contact} = \f{W}\bm{p}_{\tool,\contact} - \rot{W}{\tool}^{\contact} \, \rot{\tool}{B} \, \f{B}\bm{p}_\tool
\end{equation}
The desired orientation of the body frame $\rot{W}{B}^{\contact}$ is constructed column-wise in the tool frame $\frame{T}$ using the surface normal $\f{W}\bm{p}_{\perp,\contact}$ and the gravity vector $\bm{g}\f{W}$ and then transformed:
\begin{align}
    \tilde{\bm{\alpha}}_{R_{WT}} &= -\f{W}\bm{p}_{\perp,\contact} \times \bm{g}\f{W} \\
    \tilde{\bm{\beta}}_{R_{WT}} &= -\f{W}\bm{p}_{\perp,\contact} \\
    \bm{\alpha}_{R_{WB}} &=  \frac{1}{ \lvert \tilde{\bm{\alpha}}_{R_{WT}} \rvert} \tilde{\bm{\alpha}}_{R_{WT}} , \quad
     \bm{\beta}_{R_{WT}} =  \frac{1}{ \lvert \tilde{\bm{\beta}}_{R_{WT}} \rvert} \tilde{\bm{\beta}}_{R_{WT}} \\[10pt]
    \rot{W}{B}^{\contact} &= \begin{bmatrix}
    \bm{\beta}_{R_{WT}} \times \bm{\alpha}_{R_{WT}}, \bm{\alpha}_{R_{WT}}, \bm{\beta}_{R_{WT}}
    \end{bmatrix} \cdot \rot{T}{B}
\end{align}
To conserve right-handedness of $\rot{W}{B}^{\contact}$, the sign of the first column is flipped if $\text{det}( \rot{W}{B}^{\contact}) = -1$.

Angular and linear velocity, as well as torque is held at $0$ during contact.
\begin{equation}
    \f{W}\vec{v}_{B}^{\contact}=\vec{0},\, \f{W}\vec{\omega}_{B}^{\contact}=\vec{0},\, \f{W}\torque{B}^{\contact}=\vec{0}
    \end{equation}
    \begin{equation}
    \f{W}\force{B, \des} =  -\f{W}\bm{p}_{\perp,\contact} \, \cdot f_{\des}
\end{equation}
For the duration of the contact the force along $\vec{z}_\tool$ follows a sinusoidal ramp to the desired magnitude $f_{\des}$, is held at the desired value and ramped back to $0$.

\section{Experiments}
\label{sec:experiments}

Through a series of experiments we demonstrate the capabilities and applications of the system.
\begin{itemize}
    \item \ref{sec:experiment_disturbance}: System response to an external disturbance.
    \item \ref{sec:force_tracking}: Direct force control in interaction.
    \item \ref{sec:corner_cases}: Robustness to planner error, when a desired force is given in free space.
    \item \ref{sec:experiment_whiteboard}: Push-and-slide tracking on a planar surface, while rejecting disturbances and controlling interaction force.
    \item \ref{sec:experiment_ndt}: Viability as an infrastructure contact testing tool.
    \item \ref{sec:undulating_surface}: Statistical evaluation of the intentional interaction force control contacting an undulating structure.
\end{itemize}

State estimation for the experiments in this paper is carried out by fusing on-board \ac{IMU} data with external motion capture information from a VICON system. A video showcasing the experiments is available as supplementary material\footnote{Supplementary video link: \href{https://youtu.be/M7-cUsIyT_o}{https://youtu.be/M7-cUsIyT\_o}}.

\subsubsection{Controller Parameters}
Controller gains and parameters chosen for the experiments are shown in \cref{tab:controller_params}. 
Experimentally tuned position and velocity gains $\overline{\bm{K}}_v = \begin{bmatrix} \overline{K}_{v,\text{lin}} \eye{3} & 
\overline{K}_{v,\text{ang}} \eye{3} \end{bmatrix}^\top$ and $\overline{\bm{K}}_v = \begin{bmatrix} \overline{D}_{v,\text{lin}} \eye{3} & 
\overline{D}_{v,\text{ang}} \eye{3} \end{bmatrix}^\top$ are different for two groups of experiments due to a hardware improvement that reduced communication delay to allow for more aggressive gains. 
Several experiments used $\bm{K}_I = \begin{bmatrix} K_{I,\text{lin}} \eye{3} & K_{I,\text{ang}} \eye{3} \end{bmatrix}^\top$ of $3.0$ instead of $1.0$ to produce a more aggressively tuned wrench estimator. Effects of the wrench estimator's slow response are discussed in \cref{sec:experiment_disturbance}.

\subsubsection{Camera and Tool Frame Calibration}
 As the \ac{TOF} camera outputs intensity images on which calibration targets are detectable, we use the kalibr toolbox\footnote{\url{https://github.com/ethz-asl/kalibr}} to obtain the transformation from camera optical frame $\frame{\camera}$ to the body-fixed frame $\frame{\base}$.
 
 As the tool frame $\frame{\tool}$ changes depending on the mounted tool, its exact location w.r.t. the body-fixed frame $\frame{\base}$ also needs to be calibrated. If available, an external motion capture system can be used for this. In other cases, we perform a hand-held calibration maneuver where the end-effector is held in steady contact to a flat surface (e.g. floor) and the body is rotated in all three axes about the point of contact. Simultaneously, the current distance and surface normal in camera frame $\frame{\camera}$ are calculated from the point cloud of the \ac{TOF} camera and linear least squares batch optimization is used to obtain the contact point.
 
\begin{table}
\centering
\def\arraystretch{1.2}
\begin{threeparttable}
\centering
\begin{tabular}{c c c c c c c}
\hline
\textbf{Experiment} & $\overline{K}_{v,\text{lin}}$ & $\overline{K}_{v,\text{ang}}$ & $\overline{D}_{v,\text{lin}}$ & $\overline{D}_{v,\text{ang}}$ & $K_{f,p}$ & $K_{f,i}$ \\
\hline
A, D
& 130.0 & 9.0 & 40.0 & 3.0 & 0.1 & 1.0 \\
B, C, E, F& 100.0 & 3.5 & 35.0 & 1.2 & 0.1 & 1.0 \\
\hline
\vspace*{0.1cm}
\end{tabular}

\begin{tabular}{c c c c c c}
\hline
\textbf{Experiment} & $m^*_{v,\text{wall}}$ & $m^*_{v,\text{free}}$ & $J^*_{v}$ & $K_{I,\text{lin}}$ & $K_{I,\text{ang}}$ \\
\hline
A&
\multicolumn{3}{c}{(see \cref{tab:disturbance_test})} & 1.0 & 1.0 \\
B, C &
0.5 & 5.0 & 5.0 & 3.0 & 3.0 \\
D&
0.25 & 5.0 & 5.0 & 3.0 & 3.0 \\
E, F&
0.5 & 5.0 & 5.0 & 1.0 & 1.0 \\
\hline
\vspace*{0.1cm}
\end{tabular}

\begin{tabular}{c c c c c c}
\hline
\textbf{Experiment} & $d_{min}$ & $d_{max}$ & $e_{min}$ & $e_{max}$ & $c_{\lambda}$ \\
\hline
A, D &
0.02 & 0.2 & 0.2 & 0.25 & 0.01 \\
B, C, E, F&
0.2 & 0.4 & 0.15 & 0.25 & 0.01 \\
\hline
\end{tabular}
\caption{Controller parameters for experiments}
\label{tab:controller_params}
\begin{tablenotes}
  \item[] $^*$Inertial parameters are multipliers of the system inertia.
\end{tablenotes}
\end{threeparttable}

\vspace*{-\baselineskip}
\end{table}

\subsection{Disturbance in free flight}
\label{sec:experiment_disturbance}
In this section, we show the system response to an external disturbance when in free flight versus in undesired contact.

In order to simulate an undesired and invisible disturbance (such as a wind gust), to which the system should react with strong disturbance rejection, we perturb the system with a long, thin carbon-fiber stick. 
To mimic a desired interaction disturbance (such as wall contact), we push on the system by means of a small wooden wall mounted to the end of the stick that can be sensed by the \ac{TOF} camera.

In all experiments, the \ac{MAV} is commanded to hover at a constant position reference. The actual interaction force is measured by the force-torque sensor on the end-effector for evaluation purposes, but not known to the controller.

\begin{figure}[ht]
    \centering
    \begin{subfigure}[c]{\linewidth}
        \centering
        \includegraphics[width=0.9\linewidth]{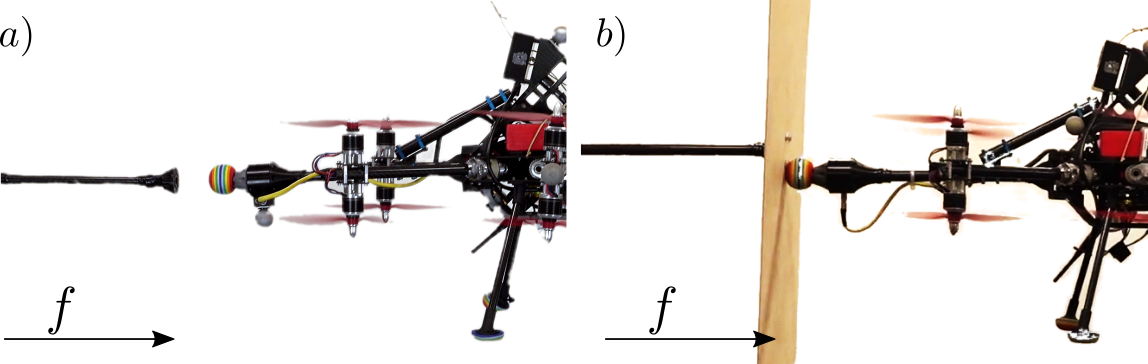}
        \subcaption{Pushing with a stick (a) and a wooden wall (b). Direction of force is indicated by the arrow $f$.}
    \end{subfigure}
    \begin{subfigure}[c]{\linewidth}
        \centering
        \includegraphics[width=\linewidth]{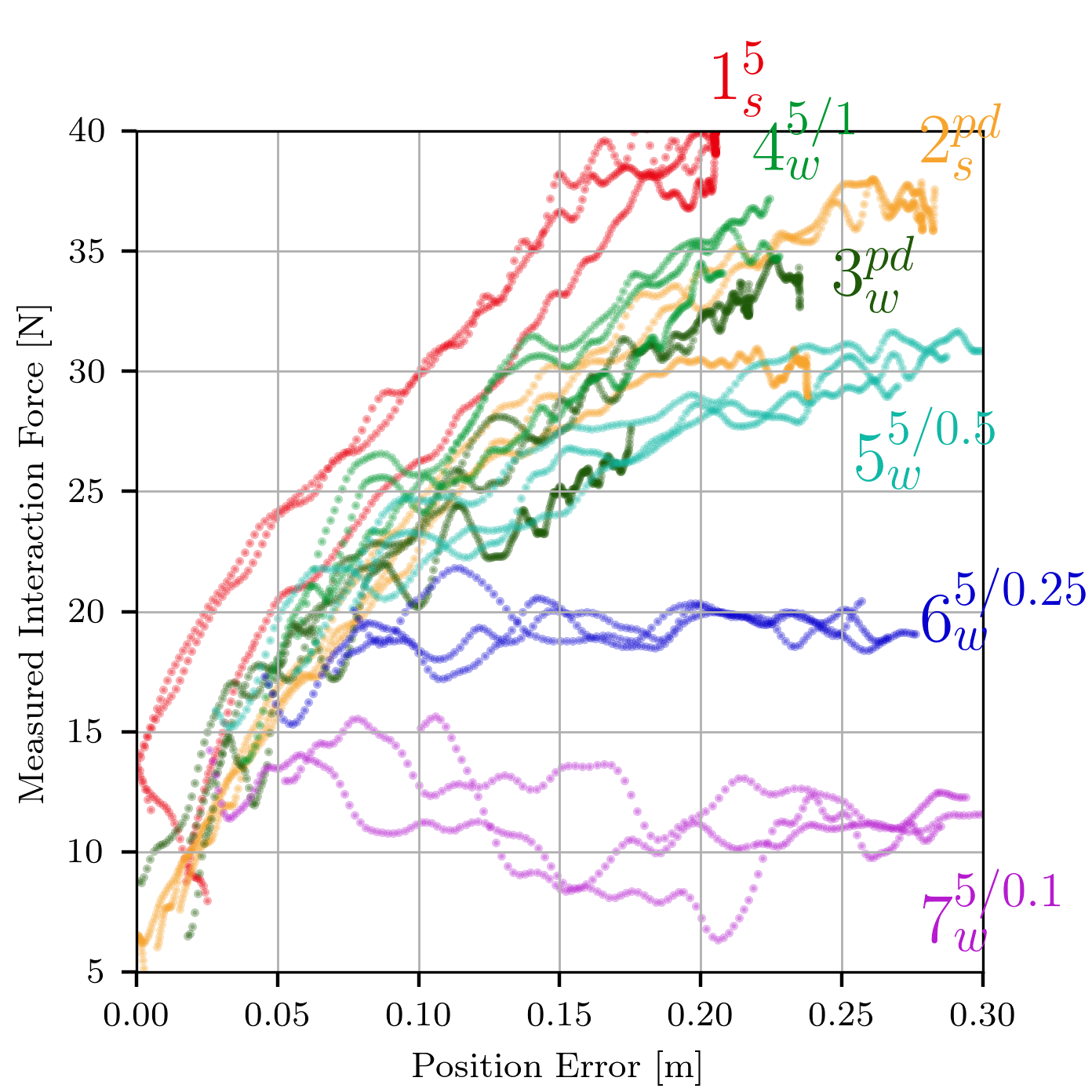}
        \subcaption{Measured force and resulting position error for 7 different controller configurations with 3 repetitions each. Horizontal lines (e.g. case $7^{5/0.1}_{w}$) indicate a very compliant response.}
    \end{subfigure}
    \caption{Free-flight disturbance experiments with a stick and a wooden wall.}
    \label{fig:disturbance_test}
\end{figure}

\Cref{fig:disturbance_test} shows the observed position error per applied force for different controller and disturbance combinations. A standard \ac{PD} controller (trials $2_{s}^{pd}$ and $3_{w}^{pd}$ shows a similar response for both disturbances. The variable \ac{ASIC} controller exhibits a very stiff behavior for high virtual mass configurations ($1_{s}^{5}$) and invisible stick disturbances. For visible wall disturbances, the observed compliance varies according to $m_\text{wall}$ (trials $4_{w}^{5/1} - 7_{w}^{5/0.1}$).
\Cref{tab:disturbance_test} gives an overview over the different parameters used, as well as the observed spring constant of the system response. The negative value of $7_{w}^{5/0.1}$ is attributed to the transient noise at the beginning of the disturbance.

Some hysteresis is apparent in the stiffness plot for the stick push with high virtual mass. This can be attributed to the relatively slow response of the momentum-based estimator with gain $\bm{K}_I = 1.0 \eye{6}$. The fidelity of the wrench estimator has a strong effect on the impedance controller's tracking ability when the system is subject to external impulses.

\begin{table}[ht]
\begin{tabular}{llllll}
\hline
                 & Type  & $m_{v,\text{free}}$ & $m_{v,\text{wall}}$  & $k\ [\si{\newton\per\meter}]$      & $\sigma_{k}\ [\si{\newton\per\meter}]$ \\ \hline
$1_{s}^{5}$      & stick &  $\boldsymbol{5.0}$ & $(0.25)$        & $135.57$ & $18.45$      \\
$2_{s}^{pd}$     & stick & N/A & N/A             & $100.25$ & $3.05$       \\
$3_{w}^{pd}$     & wall  & N/A            & N/A  & $93.46$  & $3.10$       \\
$4_{w}^{5/1}$    & wall  & $5.0$          & $\boldsymbol{1.0}$  & $92.09$  & $8.12$       \\
$5_{w}^{5/0.5}$  & wall  & $5.0$          & $\boldsymbol{0.5}$  & $47.64$  & $3.40$       \\
$6_{w}^{5/0.25}$ & wall  & $5.0$          & $\boldsymbol{0.25}$ & $4.05$   & $5.42$       \\
$7_{w}^{5/0.1}$  & wall  & $5.0$          & $\boldsymbol{0.1}$  & $-5.57$  & $0.39$      \\
\hline

\end{tabular}
    \caption{Parameters for the free-flight disturbance experiments. The observed spring-constant $k$ and its standard deviation $\sigma_{k}$ are obtained by a line fit to the 3 individual trials and quantify the stiffness of the system. }
    \label{tab:disturbance_test}
\end{table}

\subsection{Direct Force Tracking Accuracy}
\label{sec:force_tracking}
\begin{figure}[ht]
\centering
\includegraphics[width=1.0\linewidth]{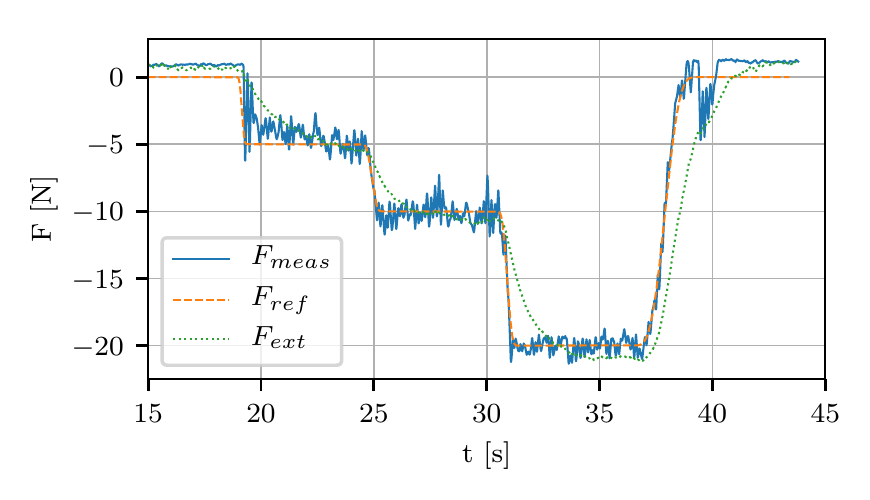}
 \caption{Tracking of 3 different force references. Dashed lines are the force references in the world frame, solid lines are measured forces at the end effector. The dotted line shows the momentum based force estimate.}
 \label{fig:force_tracking}
\end{figure}

In order to evaluate the accuracy of force tracking with \emph{intentional interaction control}, we design a trajectory containing both a pose and a force reference. The position reference is set approximately on the surface of a rigid vertical wall, and the attitude reference sets the $z_\tool$-axis orthogonal to the surface plane. The force trajectory changes between $\SI{5}{\newton}$, $\SI{10}{\newton}$, and $\SI{20}{\newton}$.
\Cref{fig:force_tracking} shows the tracking performance of the force controller. The reference force is tracked consistently, even through fast changes of the set point. The momentum based force estimate adapts more slowly to the change of force but converges to similar values.

We found that both the feed forward term and the integral gain are the most essential parts of the force controller. Increasing the proportional gain usually leads to higher frequency changes in the resulting contact forces, while not significantly improving the response time to reference changes.

\begin{figure}[ht]
\centering
\begin{subfigure}[c]{\linewidth}
\centering
\includegraphics[width=\linewidth]{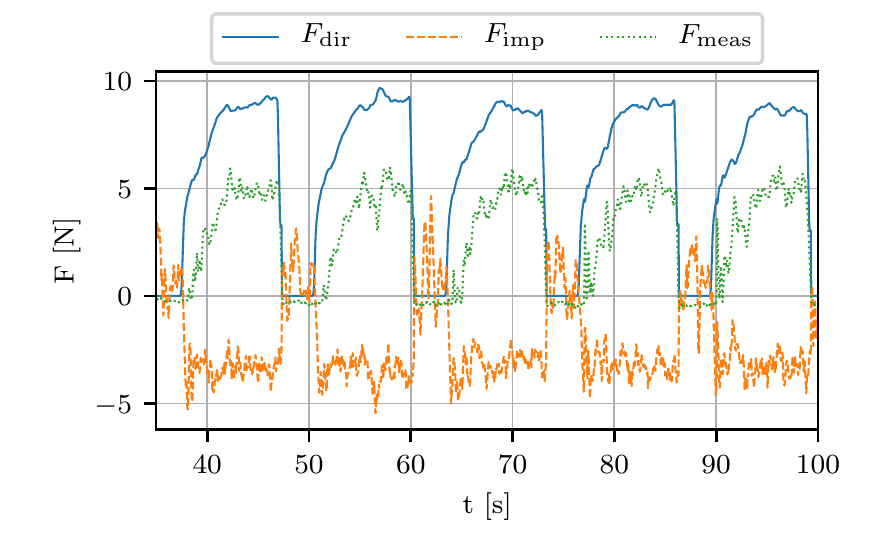}
\vspace{-0.5cm}
 \caption{Direct force control, impedance control, and measured force during contact. The position reference is $\SI{4}{\centi\meter}$ in front of the wall. Direct force control compensates for position tracking of impedance control, for a reference force of $\SI{5}{\newton}$.}\label{fig:force_impedance_measured}
 \end{subfigure}
 \begin{subfigure}[c]{\linewidth}
 \centering
\includegraphics[width=\linewidth]{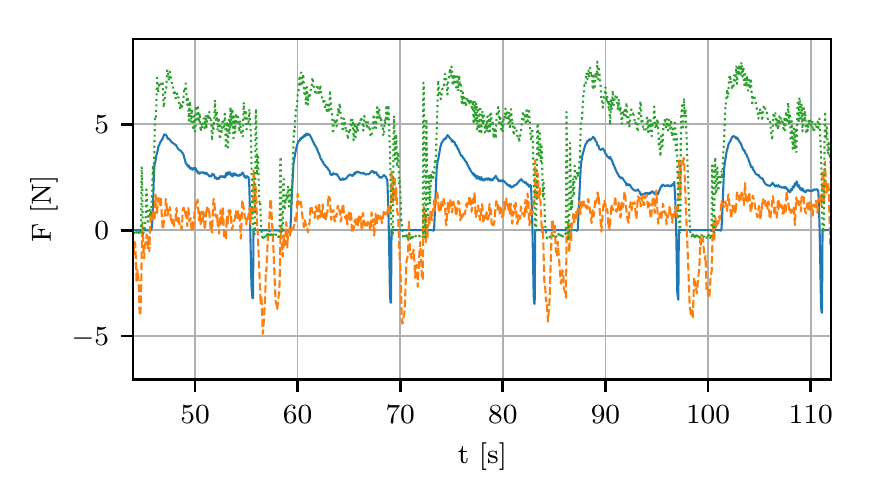}
\vspace{-0.5cm}
 \caption{Contact inspection with position reference $\SI{1}{\centi\meter}$ behind the wall. Impedance and direct force control components complement to track the reference of $\SI{5}{\newton}$.}
 \label{fig:force_impedance_measured2}
 \end{subfigure}
 \caption{Comparison of the controlled force resulting from direct force and impedance control components.}
\end{figure}

Additionally, we test the robustness of the proposed \emph{intentional interaction control} to small errors in the planned trajectory by performing a surface inspection task with multiple contacts with reference forces of $\SI{5}{\newton}$. In a first experiment, we set the position reference $\SI{4}{\centi\meter}$ in front of the true surface position, and in a second trial we set it $\SI{1}{\centi\meter}$ behind the true position. Plots of the body force command in the contact direction for the two trials are shown respectively in \cref{fig:force_impedance_measured} and \cref{fig:force_impedance_measured2}, also depicting the interplay of force and impedance control components as described in \cref{eq:combined_command}, and shown as a block diagram in \cref{fig:force_control_detail}. The figures illustrate that for each trial, the measured force settles at the reference force after a short response time. The impedance control component holds a constant offset term based on the position error when the end effector is against the contact surface. The direct force control term compensates for this error, driving the force error to zero primarily with the integral term. The feed forward force control term initially causes an undershoot in force (\cref{fig:force_impedance_measured}) where the position set point is away from the wall, and an overshoot in force (\cref{fig:force_impedance_measured2}) where the set point is into the wall.
The experiments show that the direct force control command $\force{\text{dir}}$ compensates the impedance control command $\force{\text{imp}}$ in order to achieve the desired contact force.

A measured force offset in free flight in \cref{fig:force_tracking} highlights that force sensors are not immune to error and offsets. Cooling effects of the propellers were not compensated for in these tests, and resulted in a slowly changing bias. The resulting performance still demonstrates an ability to track the quick response of an integrated force sensor.

\subsection{Force Tracking Robustness to Planner Error}
\label{sec:corner_cases}
\begin{figure}[ht]
\centering
\begin{subfigure}[c]{\linewidth}
\centering
\includegraphics[width=0.9\linewidth]{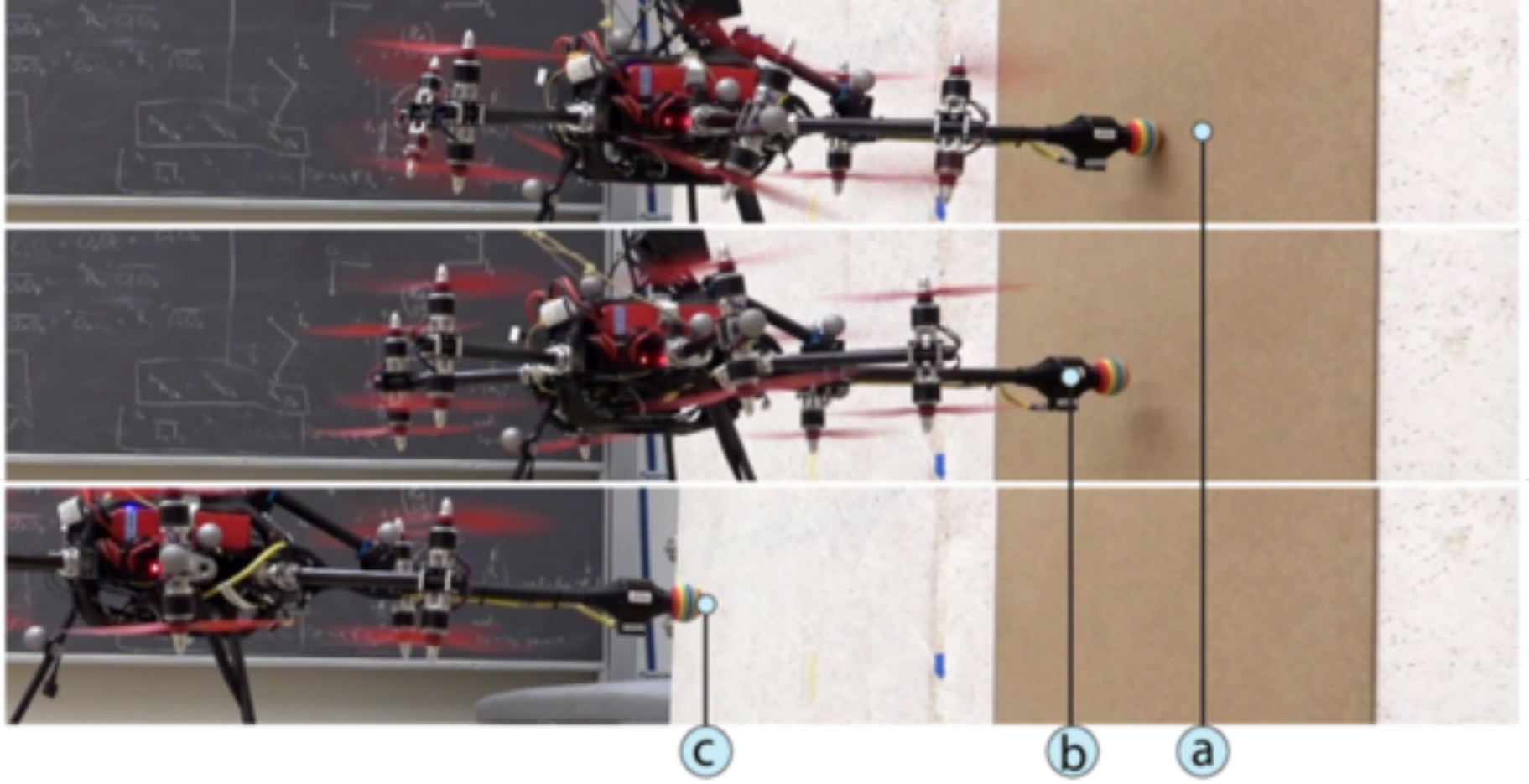}
\label{fig:picture_cornerCases}
 \end{subfigure}
 
 \begin{subfigure}[c]{\linewidth}
\centering
\includegraphics[width=0.9\linewidth]{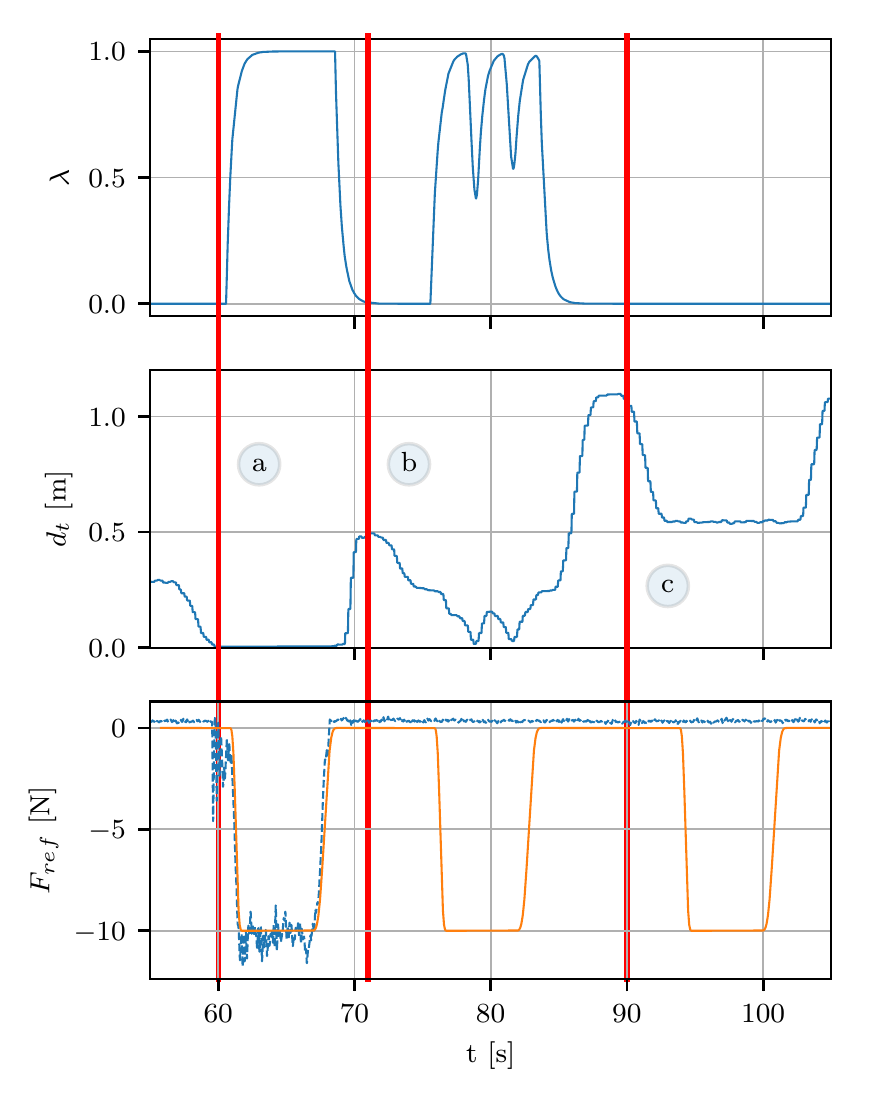}
 
 \label{fig:lambda_wallDist}
 \end{subfigure}

 \caption{Confidence factor $\lambda$, wall distance and force reference show behavior for a set point a) behind the wall, b) in front of the wall by $\SI{0.25}{\meter}$, and c) $>\SI{0.5}{\meter}$ away from the wall. }
   \label{fig:cornerCases}

\end{figure}

To evaluate the robustness of intentional interaction control from \cref{sec:intentional_interaction} to planner error, we specify a desired force in three situations that are not at a contact surface:
\begin{enumerate}[label=(\alph*)]
    \item Close to a surface
    \item \SI{0.25}{\meter} in front of the surface
    \item \SI{0.5}{\meter} away from the surface
\end{enumerate}
\Cref{fig:cornerCases} shows the resulting behavior for the three scenarios. In a), the platform is close enough to the wall to enable direct force control by increasing the confidence factor $\lambda$ to 1.
In b), the perceived tool distance $d_t$ in combination with the tool error $\bm{e}_t$ leads to a short increase in $\lambda$ before being pulled back by the resetting force of the impedance controller.
In c), the tool distance $d_t$ is larger than the maximum selected tool distance $d_{\text{max}}$ and direct force control is therefore not enabled.
In all cases, the system responds to planner error in a stable way, and is able to continue executing a compromise of the combined state and force trajectory.

The behavior of each case depends on chosen values for $d_{\text{min}}$, $d_{\text{max}}$, $e_{\text{min}}$ and $e_{\text{max}}$, and how they generate $\lambda$ as described in \cref{eq:lambda}. Case a) and c) are desired effects when the system is close enough that establishing contact is more important than position tracking, and when the system is far enough from a surface that force tracking should not even be attempted. In case b), which is visualized in \cref{fig:lambda}, the contact surface is detected but the position error is too large to track the tool tip at the surface. This case can be avoided by choosing $d_{\text{max}} \leq e_{\text{min}}$. Provided that nominal tracking error remains small, the system remains in a position tracking region until the detected surface is within the tracking error margin.

\subsection{Push-and-Slide Along a Flat Surface}
\label{sec:experiment_whiteboard}

We evaluate the system's ability to reject disturbances from friction, while accurately and repeatably drawing on a whiteboard positioned in a known location, for both direct force and for impedance control. The trajectory traces a spline with the tool point on the surface of the whiteboard. The end-effector is a whiteboard marker with no additional compliance.

\begin{figure}
\centering
\includegraphics[width=0.9\linewidth]{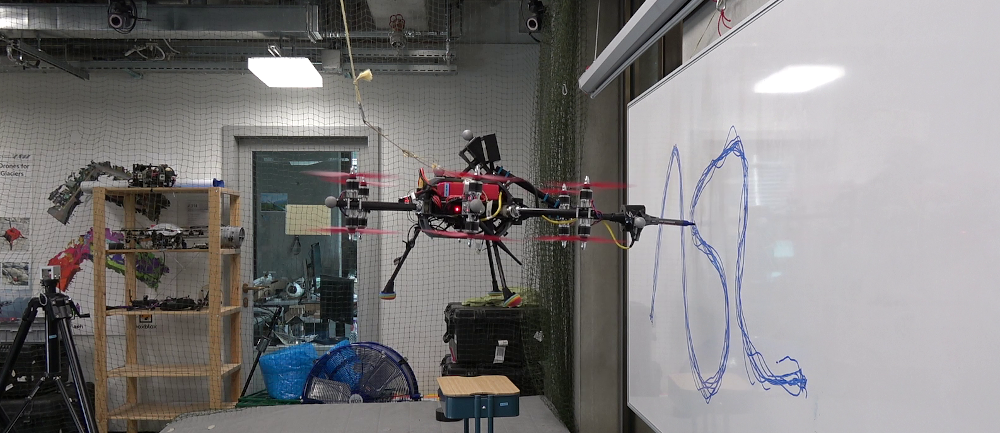}
\hspace{0mm}
\includegraphics[trim=0cm 0cm 0cm 0cm, width=0.45\textwidth]{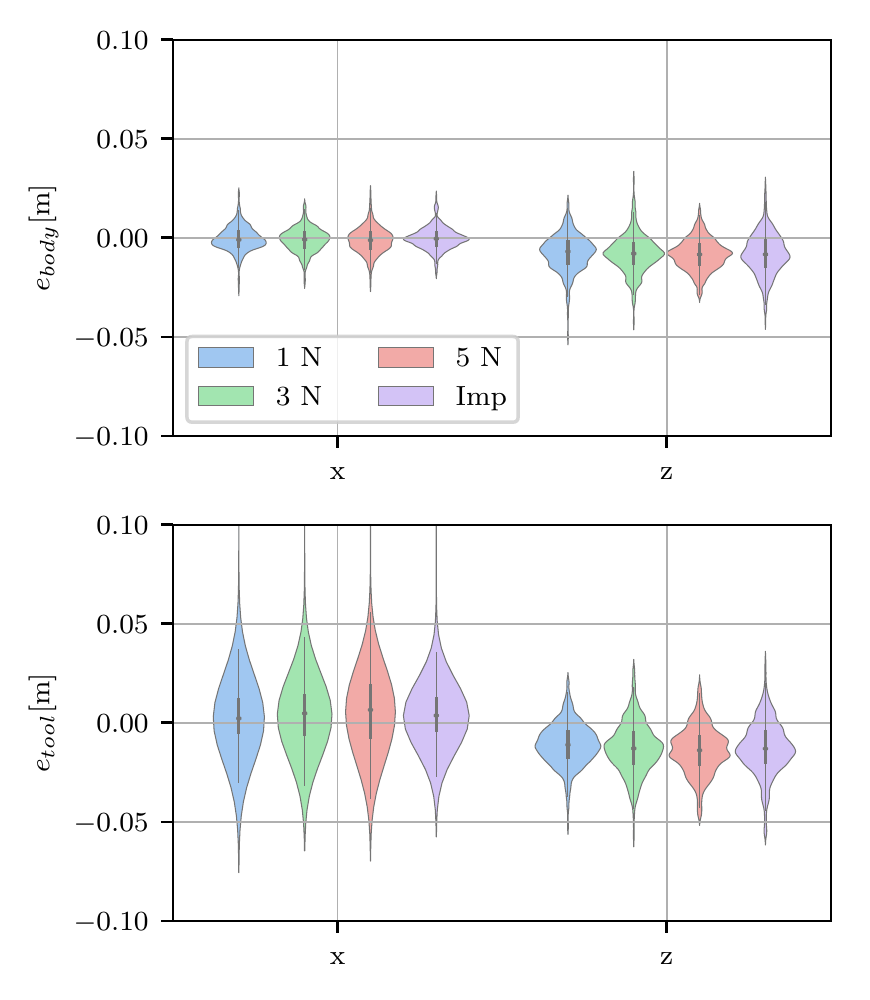}
\caption{Body (upper) and tool (lower) position tracking accuracy in the inertial frame as violin plots for push-and-slide experiments. 3 trials of each test are represented in the data, for reference force tracking of 1, 3, and 5 \si{\newton} and impedance control.}
\label{fig:whiteboard}
\end{figure}

\begin{figure}
\centering
\includegraphics[width=0.9\linewidth]{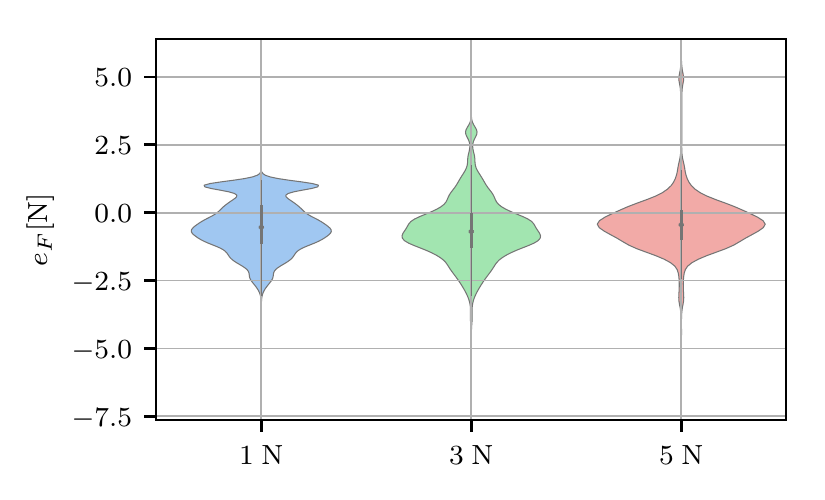}
\caption{Force tracking accuracy for push-and-slide, using 3 different reference forces and repeating the trajectory three times for each reference force. The violin plots show the errors between the commanded and the measured normal force on the whiteboard, once contact has been established.}
\label{fig:whiteboard_force}
\end{figure}

For direct force control, the force trajectories are designed to smoothly ramp up to the desired force reference, once the position reference has reached the whiteboard surface. We compare the tracking performance of three different force references of \SI{1}{\newton}, \SI{3}{\newton}, and \SI{5}{\newton}. The controller parameters are listed in \cref{tab:controller_params}.
All experiments are performed three times to allow for a statistical analysis. \Cref{tab:whiteboard_numerical_results} and \cref{fig:whiteboard} to \cref{fig:whiteboard_force} show the position and force errors of the experiments. As the whiteboard is aligned with the inertial $x$-$z$-plane, position errors are only evaluated on this plane.

\begin{table}[ht]
\centering
\begin{tabular}{llllll}
\hline
 Type  & $e_{B,x}$ & $e_{B,z}$  & $e_{t,x}$ & $e_{t,z}$  & $e_{f}$\\
 \hline
Impedance & 0.0055 & 0.0131 & 0.0962 & 0.0183 & - \\
1 N & 0.0063 & 0.0118 & 0.1319 & 0.0152 & 1.0627 \\
3 N & 0.0061 & 0.0124 & 0.1096 & 0.0180 & 1.2761 \\
5 N & 0.0067 & 0.0115 & 0.1049 & 0.0181 & 1.0963 \\
\hline

\end{tabular}
    \caption{Numerical results for push-and-slide experiments. Printed are RMSE for the body and the tool position (in \si{\meter}) as well as for force tracking (in \si{\newton}).}
    \label{tab:whiteboard_numerical_results}
\end{table}

Without any change in the controller, the system is able to handle transitions in and out of contact with good stability, and without significant tracking error on the surface plane.
The system demonstrates rejection of torque and lateral force disturbances caused by surface friction while maintaining a consistent contact force against the wall.

\Cref{fig:whiteboard} shows that the body reference position is tracked equally well for the 4 different cases, while the tool position exhibits a larger tracking error. We attribute this to the fact that the controller has not been designed to track a position reference at the end-effector, but rather at the body center. Despite using high stiffness parameters for attitude and lateral position control, unmodeled effects such as lateral friction and stiction at the whiteboard, as well as a larger tool mass due to the force sensor, prevent more accurate end-effector tracking.

The force tracking results in \cref{fig:whiteboard_force} show similar error for all three reference force magnitudes.

\subsection{Potential Field Concrete Inspection Task}
\label{sec:experiment_ndt}
Similar to experiments in \cite{bodie2019omnidirectional}, we conduct an autonomous contact inspection task on a sample of reinforced concrete to demonstrate that an autonomous \ac{MAV} can achieve comparable results to standard measurements taken by hand.

We equip the end effector with an \ac{NDT} contact sensor that measures both the electrical potential difference between a saturated Copper Sulfate Electrode (CSE) and the embedded steel, and the electrical resistance between the sensor on the concrete surface and the steel reinforcement. 
Electrical potential and resistance results can be used as an indicator for the corrosion state of the steel \cite{bertolini2013corrosion}. A cable is connected to the reinforcement in the concrete structure, and is physically routed to the sensor on the flying system to perform the measurements.
The concrete specimen used for this experiment has a known corrosion spot at a certain location and a constant cover depth. The corrosion state can therefore be evaluated against this information.
The concrete block is positioned at a known location, and a trajectory is defined to contact 9 points at \SI{5}{\cm} intervals along the surface.
Each point is held for a duration of \SI{10}{\second}, during which a reference force of \SI{5}{\newton} is requested.

\begin{figure}[ht]
\centering
\begin{subfigure}[c]{\linewidth}
\includegraphics[width=1.0\linewidth]{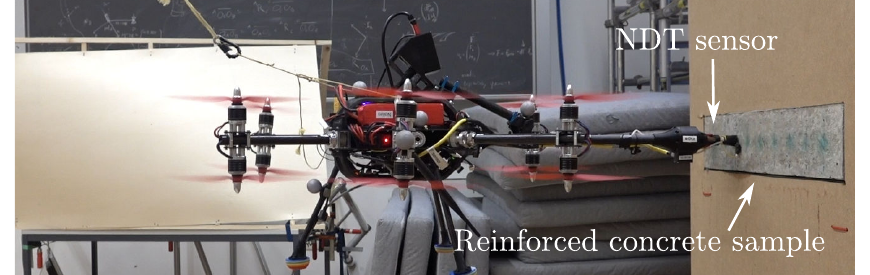}
 \label{fig:concreteSample}
 \end{subfigure}
 
\begin{subfigure}[c]{\linewidth}
\includegraphics[width=1.0\linewidth]{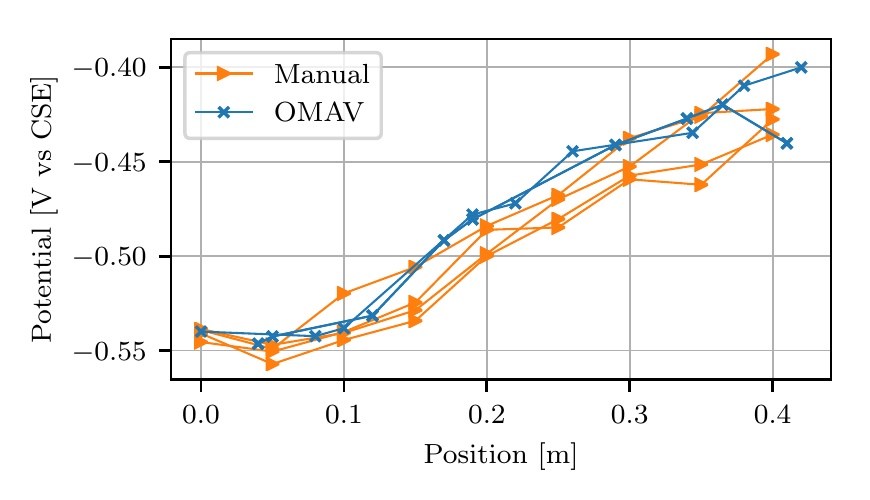}

 \label{fig:potential_2d}
 \end{subfigure}
 \caption{Potential field concrete inspection experiment. Top: System in contact with one of 9 sampling points. Bottom: Comparison of potential mapping results for measurements taken during two autonomous \ac{MAV} flights, and four manual measurements along a reinforced concrete block sample. }
 \label{fig:potential_concrete_sampling}
\end{figure}

\Cref{fig:potential_concrete_sampling} compares the autonomously measured potentials of two flights with manually measured potentials along the sample, measured before and after the flights. 
The plot shows that the controller is able to hold contact between the end effector and the sample to allow accurate measurements of the potential.

\subsection{Statistical Validation on an Undulating Surface}
\label{sec:undulating_surface}
We perform a statistical evaluation of the intentional interaction control to validate its repeatability and performance. The repeated random experiments aim to characterize the system in a more diverse, less controlled environment in order to show its applicability to complex tasks.

\begin{figure}[ht]
    \centering
    \begin{subfigure}[c]{0.49\columnwidth}
        \centering
        \includegraphics[width=0.75\linewidth]{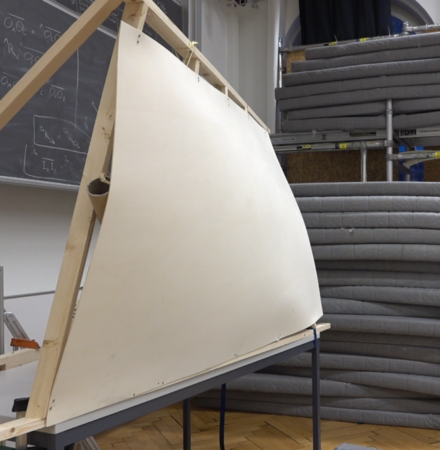}
    \end{subfigure}
    \hfill
    \begin{subfigure}[c]{0.49\columnwidth}
        \centering
        \includegraphics[width=1.0\linewidth]{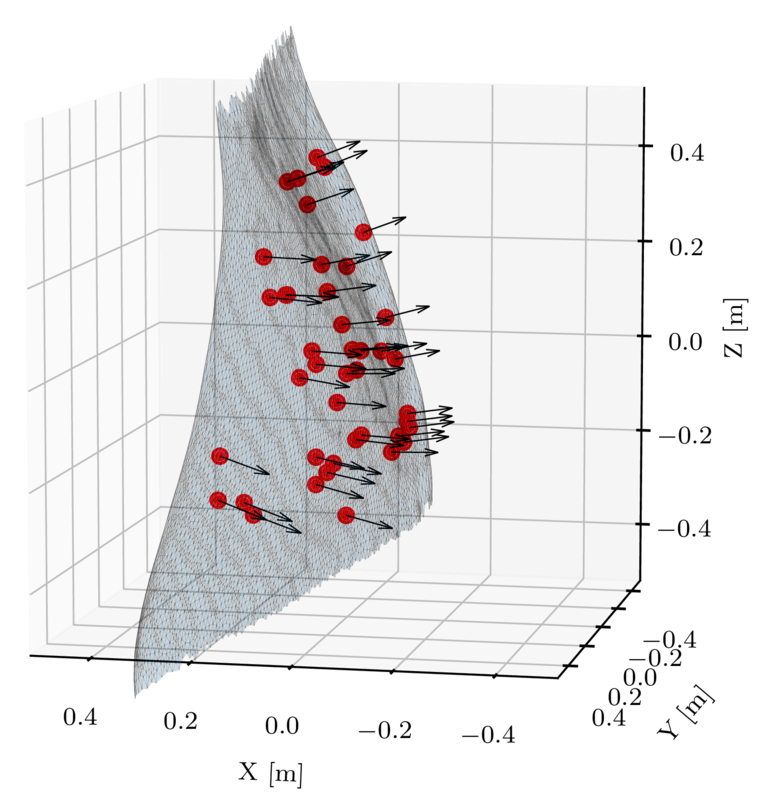}
    \end{subfigure}
    \caption{Undulating wooden wall used for experiments (left), randomly sampled contact locations (red dots) and their surface normal (black arrows).}
    \label{fig:wall_description}
\end{figure}
As a test surface, we use a doubly curved wooden surface with a size of approx. $\SI{1}{\meter} \times \SI{1.8}{\meter}$, which is mapped and used for planning as described in \cref{sec:meshplanning}. \Cref{fig:wall_description} visualizes the randomly selected $42$ contact locations at which we command the \ac{MAV} to exert a force of $\SI{10}{\newton}$ perpendicular to the surface for $\SI{5}{\second}$.

\begin{figure}[ht]
    \centering

    \begin{subfigure}{\columnwidth}
        \includegraphics[width=1.0\linewidth]{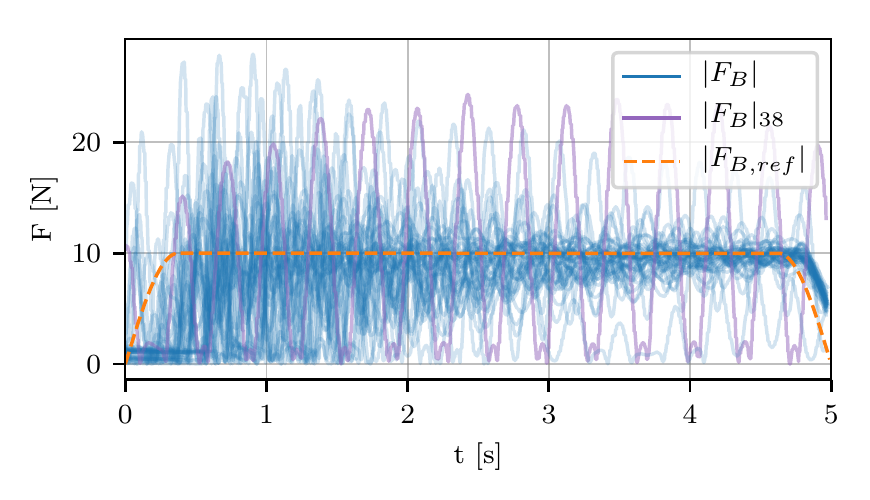}
    \centering
    \end{subfigure}
    \begin{subfigure}{\columnwidth}
        \includegraphics[width=1.0\linewidth]{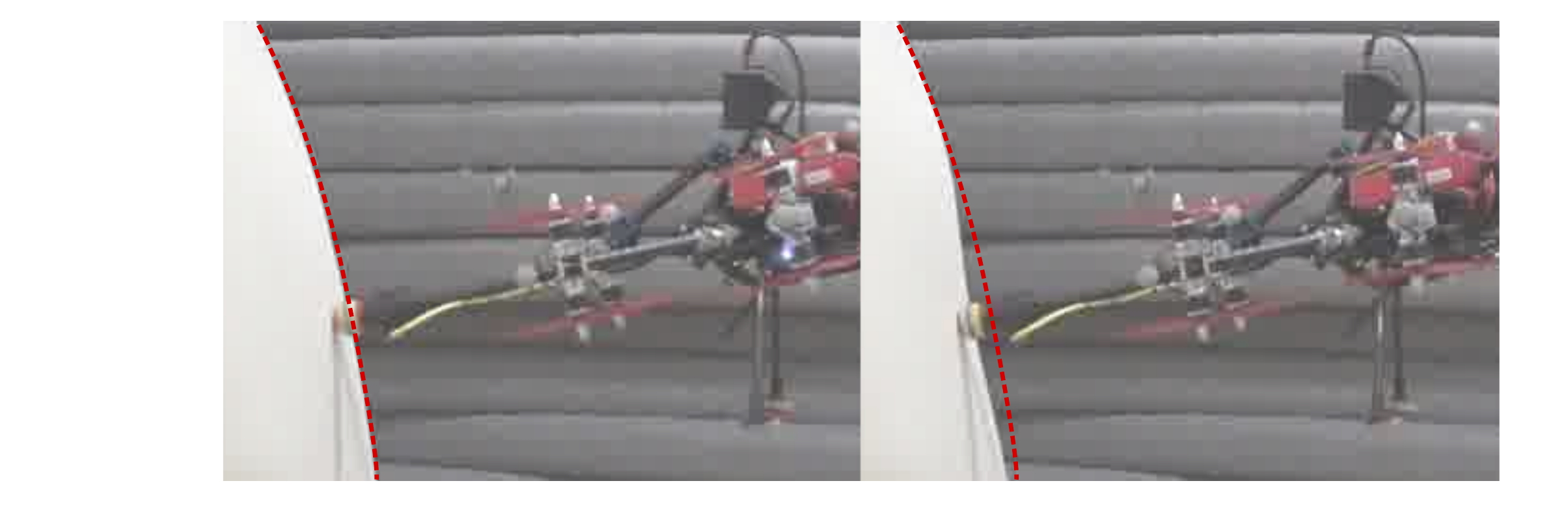}
    \centering
    \end{subfigure}
    \caption{Top: Overlay of interaction force magnitude during $42$ individual trials. Time $t=0$ corresponds to the first force command issued. Trial $38$ is marked separately, as the force did not stabilize. Bottom: Frames $\SI{0.15}{s}$ apart during Trial $38$. Deformation of the surface is clearly visible. The red dotted line indicates the nominal surface.}
    \label{fig:interaction_force}
\end{figure}

\Cref{fig:interaction_force} overlays the force along the $z_{T}$ axis for all trials with the time aligned beginning from the first force command. In all $42$ trials the system was able to hold contact and keep position and orientation regardless of the local surface geometry. The desired force of $\SI{10}{\newton}$ could be achieved in almost all trials. However, the stabilization time of the force magnitude is dependent on the local springiness of the surface. In extreme cases, such as trial $38$, the surface deflected by multiple centimeters as visualized in \cref{fig:interaction_force}.

\begin{figure}[ht]
    \centering
    \includegraphics[width=1.0\linewidth]{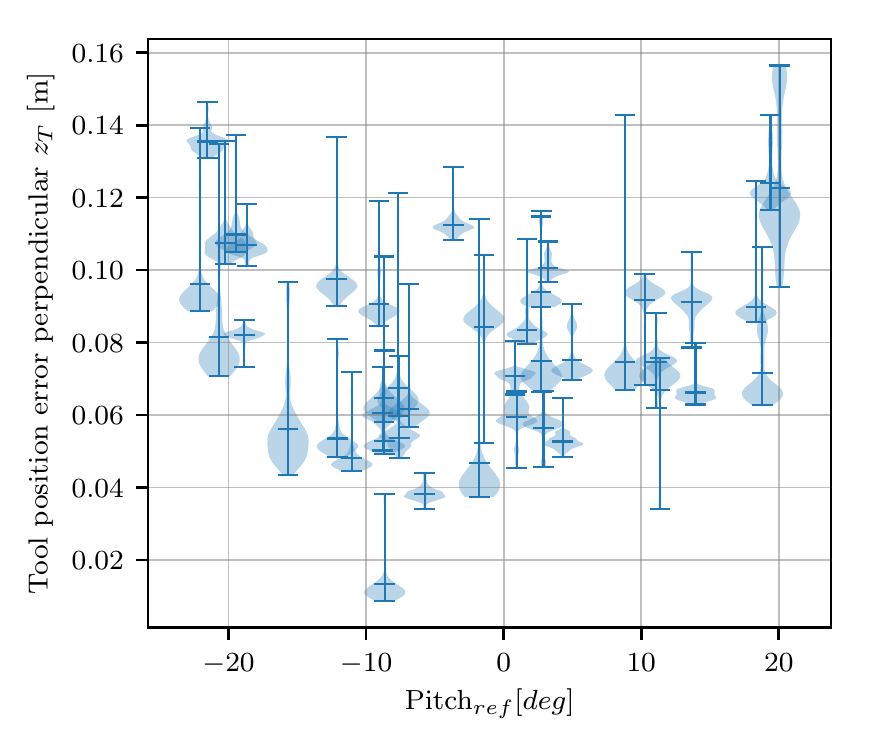}
    \caption{Tool position error perpendicular to the $z_\tool$-axis for each individual trial. Presented data is for each trial whenever the tool is within \SI{5}{\centi\meter} of the surface. Data is truncated such that each trial includes $1250$ measurements, about \SI{5}{\second} of data.}
    \label{fig:statistics_pos}
\end{figure}

An important application-driven evaluation is the $2d$ tool position error on the surface, which combines the effect of the body position and attitude error.
\Cref{fig:statistics_pos} visualizes the statistics of the tool position error for each individual trial as violin plots. The trials are sorted by their pitch reference angle, where negative pitch indicates a downward pitch of the \ac{MAV}.
Note that the tool position does not change after contact is made steadily, thus the plot in \cref{fig:statistics_pos} uses data obtained whenever the tool is closer than 5 cm to the surface.

\begin{figure}[ht]
    \centering
    \includegraphics[width=1.0\linewidth]{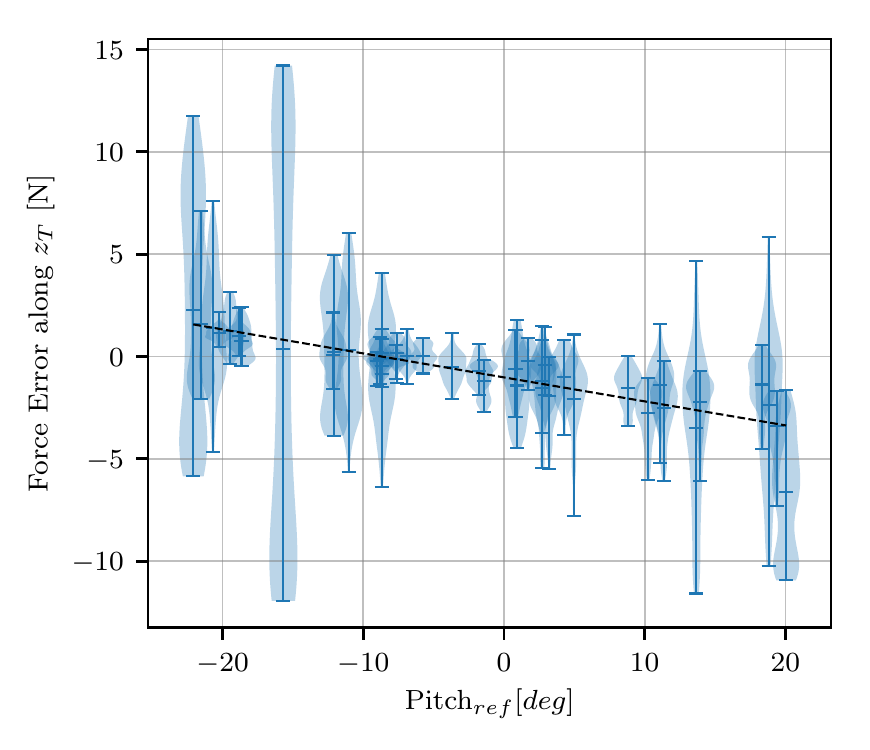}
    \caption{Force control error vs. desired pitch for each individual trial.
    Force data is truncated to the last \SI{2.5}{\second} of the interaction duration, allowing the system to stabilize first.
    The black dotted curve is a linear fit to the mean force error.}
    \label{fig:statistics_force}
\end{figure}

Similarly, \cref{fig:statistics_force} shows the force error along $z_{T}$ for each individual trial. The two curves at around $-17$ and $-22$ degrees pitch that show large densities close to the extreme values correspond to the before discussed cases where the surface springiness leads to force oscillation.

Both the position and force error statistics indicate a slight influence of the pitch reference on the errors. The system slightly overshoots the desired force if it pitches downwards and vice-versa. Overall, the system showed predictable and robust force trajectory executing during all trials.

\section{Discussion}
\label{sec:discussion}

The experiments show that selective impedance control presents a suitable method for contact inspection, if the axis of desired pressure on the surface is known. Using low virtual mass along this axis allows compliance, while high virtual mass along the axes parallel to the surface yield accuracy in lateral positioning. Further integrating variable \ac{ASIC} with distance sensing balances compliance in interaction with strong free flight disturbance rejection.

If specific interaction forces are required and if the location of the surface is only known with a precision of a few centimeters, variable \ac{ASIC} in combination with direct force control proved to give good force tracking accuracy. For position references behind or in front of the actual surface, the reference force was achieved in most experiments after few seconds of contact.
The use of the confidence factor $\lambda$ allows simple and smooth transitioning between different stages of flight, which yielded stable and controlled maneuvers while switching from free flight to interaction control.

Regardless of the control, the system's flight performance is considerably sensitive to its hardware calibration. Small changes of the \ac{COM} or slightly incorrect zeroing of tilt arms can lead to different behavior and might require new calibration. This model error is amplified by unmodeled effects, such as airflow interference or backlash in gears that are driving the tilt arms.
We aim to mitigate calibration errors in the future with automatic on-line calibration of relevant physical system properties, such as \ac{COM}, inertia and tool mass.

Finally, the experiments confirm that the presented control approach combining distance sensing and surface-based force trajectory planning complete the basis for high-level force interaction tasks to be carried out by fully actuated \acp{MAV}.

\section{Conclusion}
\label{sec:conclusion}
In this paper, we presented active interaction force control of a fully actuated \ac{MAV}. By using additional sensing such as force/torque and surface distance measurements, our system is able to reliably and safely perform force control in a variety of environments.

Extensive experiments demonstrated the feasibility of the proposed system for applications such as non-destructive testing of infrastructure and other contact-based application.

\section*{Acknowledgment}
This work was supported by funding from ETH Research Grants, the National Center of Competence in Research (NCCR) on Digital Fabrication, NCCR Robotics, and Armasuisse Science and Technology. 

\bibliographystyle{IEEEtran}
\bibliography{references}
\vspace{-1.25 cm}
\begin{IEEEbiography}
 [{\includegraphics[width=1in,height=1.25in,clip,keepaspectratio]{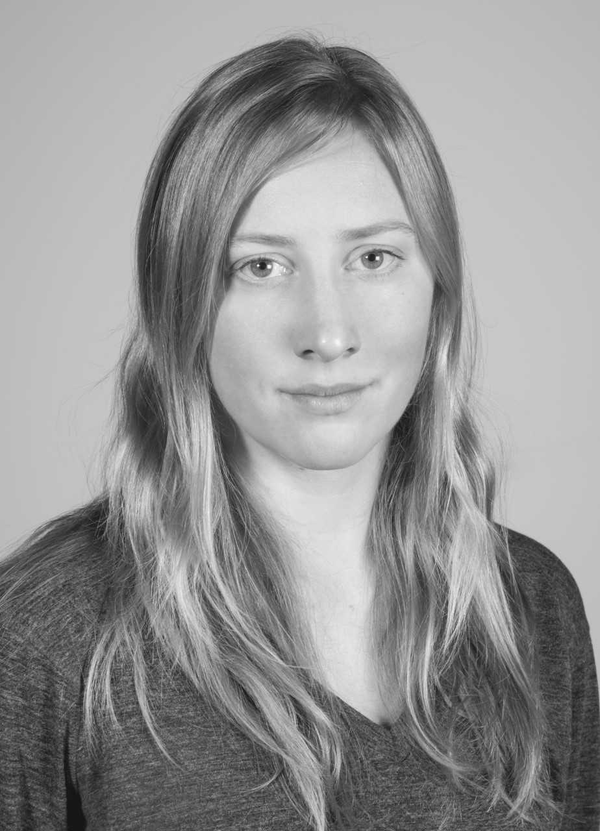}}]{Karen Bodie} received a B.S. degree in mechanical engineering from the McGill University, Montreal, Canada, in 2011, and a M.Sc. degree in mechanical engineering in 2016 from ETH Zurich, Switzerland. She is currently working toward a Ph.D. degree in mechanical engineering at the Autonomous Systems Lab at ETH Zurich. 
Her areas of research include the design and control of fully actuated aerial manipulators with the goal of high performance aerial physical interaction.

\end{IEEEbiography}
\vspace{-1.25 cm}
\begin{IEEEbiography}
[{\includegraphics[width=1in,height=1.25in,clip,keepaspectratio]{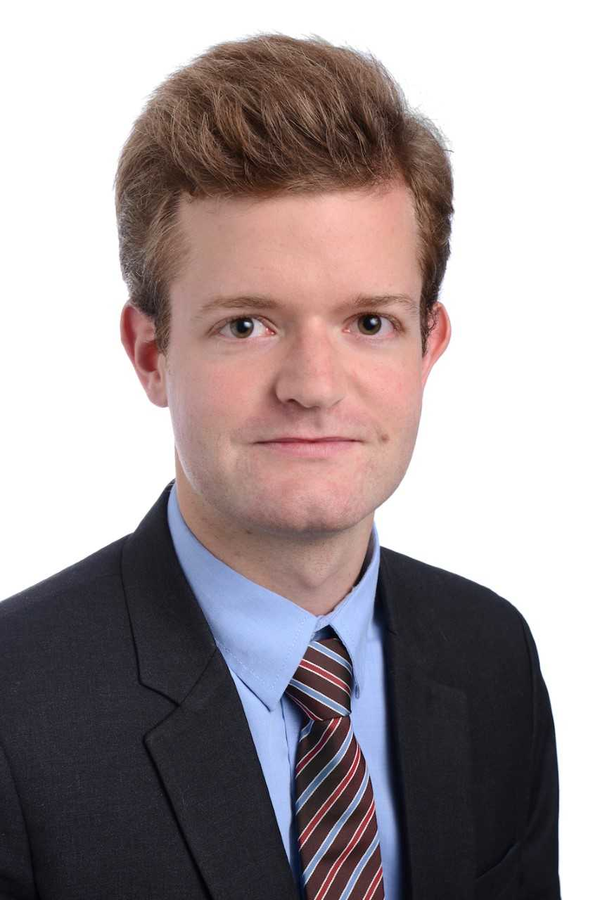}}]{Maximilian Brunner}
received his M.Sc. degree in mechanical engineering from ETH Zurich in 2017. During his studies, he focused on optimal control and path planning of both aerial and ground vehicles, and spent one year working in automotive and aerospace industry. He is now pursuing the Ph.D. degree in mechanical engineering at the Autonomous Systems Lab at ETH Zurich, where he is working on the control of overactuated multicopters, especially in the field of physical interaction with their environment.
\end{IEEEbiography}
\vspace{-1.25 cm}
\begin{IEEEbiography}
 [{\includegraphics[width=1in,height=1.25in,clip,keepaspectratio]{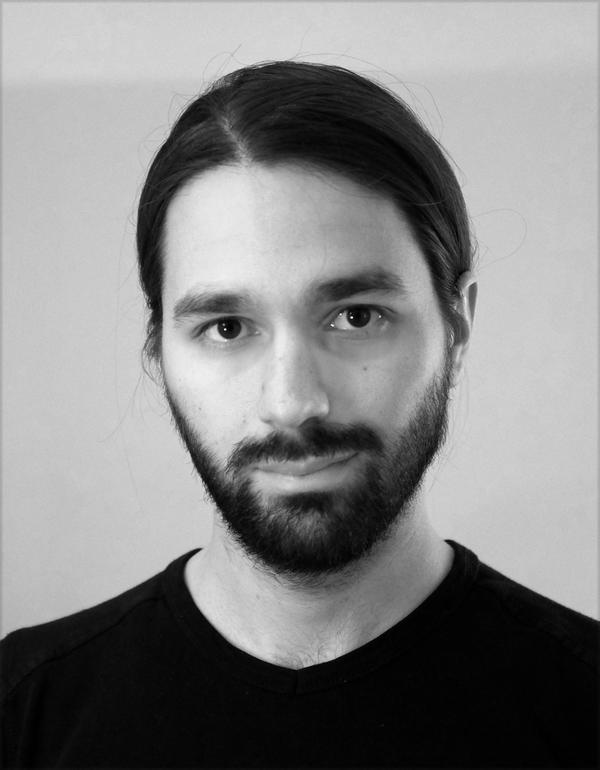}}]{Michael Pantic} received a B.Sc. degree in software engineering from Lucerne University of Applied Sciences and Arts, Switzerland, in 2013. From 2008-2016 he held various positions in industry in software engineering, architecture and consultancy. He received a M.Sc. degree in Robotics, Systems and Control from ETH Zurich in 2018, with a 9 month stay at EPF Lausanne. Currently he is pursuing a Ph.D. degree at the Autonomous Systems Lab at ETH Zurich. His area of research includes perception and planning for high-accuracy physical interaction in large workspaces with aerial manipulators.
\end{IEEEbiography}
\vspace{-1.25 cm}
\begin{IEEEbiography}
 [{\includegraphics[width=1in,height=1.25in,clip,keepaspectratio]{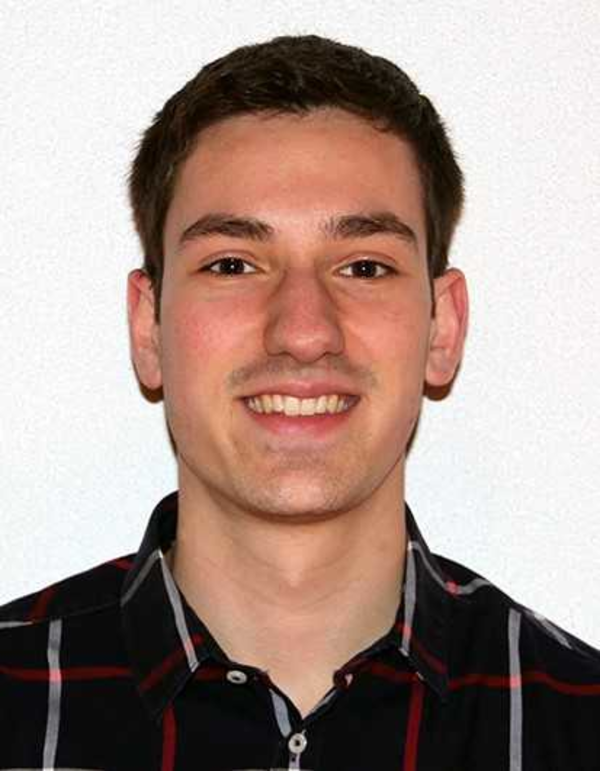}}]{Stefan Walser}
 received his bachelor’s degree in electrical engineering and his master’s degree in robotics, systems and control from ETH Zurich, Switzerland in 2018 and 2020, respectively. From 2017 to 2018 he completed an internship in an advanced robotics laboratory in Eschen, Liechtenstein. During his master's studies he did research on control algorithms for aerial vehicles with traditional approaches and machine learning at the Autonomous Systems Lab of ETH Zurich. Currently he is working as a software engineer in Zurich, Switzerland.
\end{IEEEbiography}
\vspace{-1.25 cm}
\begin{IEEEbiography}
[{\includegraphics[width=1in,height=1.25in,clip,keepaspectratio]{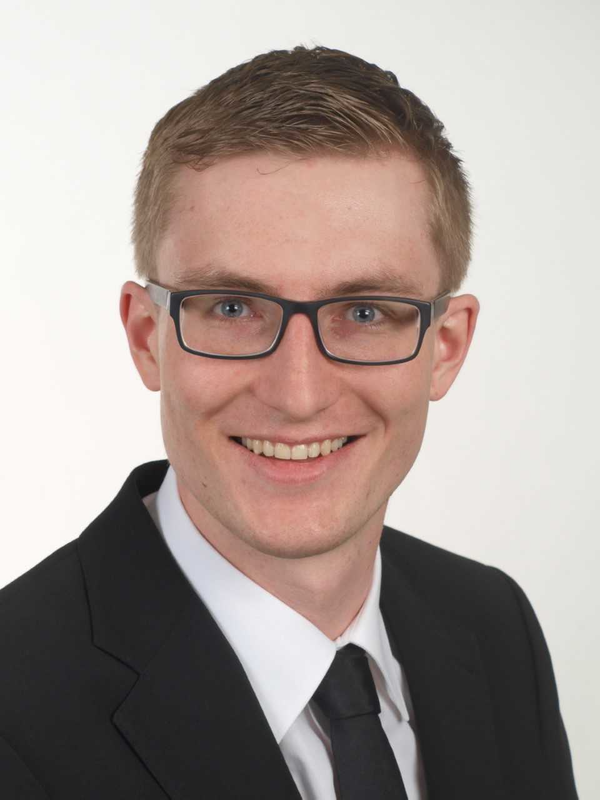}}]{Patrick Pf\"{a}ndler}
received his degrees in civil engineering in 2015 (BSc) and the Master’s degree in 2017 from ETH Zurich in Switzerland. His specialization during the Master degree was Structural Engineering and Material and Mechanics. He is currently working towards the PhD degree at the Institute of Building Materials at ETH Zurich. His research interests include the corrosion assessment of reinforced concrete infrastructures by applying well-established non-destructive testing methods and by on-site automation with robots.
\end{IEEEbiography}
\vspace{-1.25 cm}
\begin{IEEEbiography}
[{\includegraphics[width=1in,height=1.25in,clip,keepaspectratio]{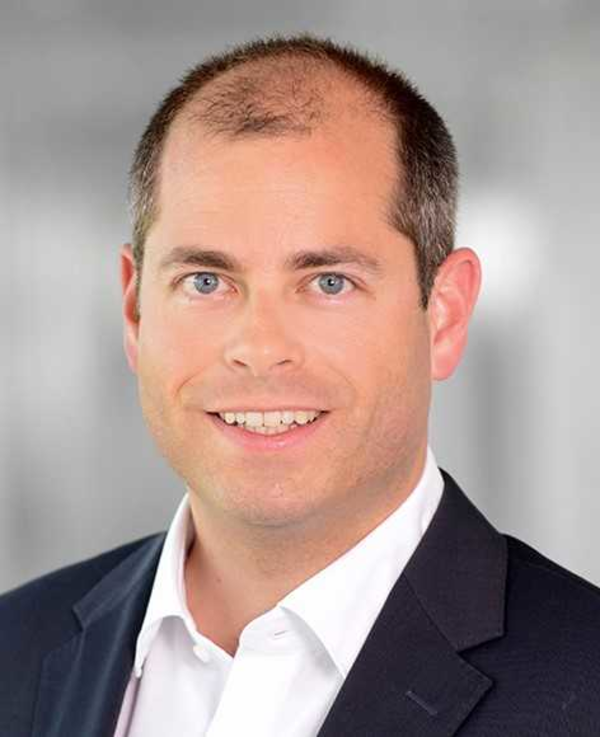}}]{Ueli Angst}
obtained his degrees in civil engineering from ETH Zurich in Switzerland (MSc) and from the Norwegian University of Science and Technology, NTNU, in Trondheim, Norway (PhD, 2011). From 2011 to 2016, he held a part-time position as Postdoc at the Institute for Building Materials at ETH Zurich, and simultaneously a part-time position as a corrosion consultant at the Swiss Society for Corrosion Protection, which is the leading agency in the field of corrosion in Switzerland. Since January 2017, Ueli Angst is an assistant professor at ETH Zurich. His research group uses experimental and computational methods covering corrosion science, electrochemistry, materials science, porous media and reactive mass transport, and civil engineering. Prof. Angst is committed to provide mechanistic insight into corrosion mechanisms and its effects on structural behavior, develop methods and sensors for monitoring purposes, robot-assisted inspection methods, and corrosion mitigation strategies.
\end{IEEEbiography}
\vspace{-1.25 cm}
\begin{IEEEbiography}
[{\includegraphics[width=1in,height=1.25in,clip,keepaspectratio]{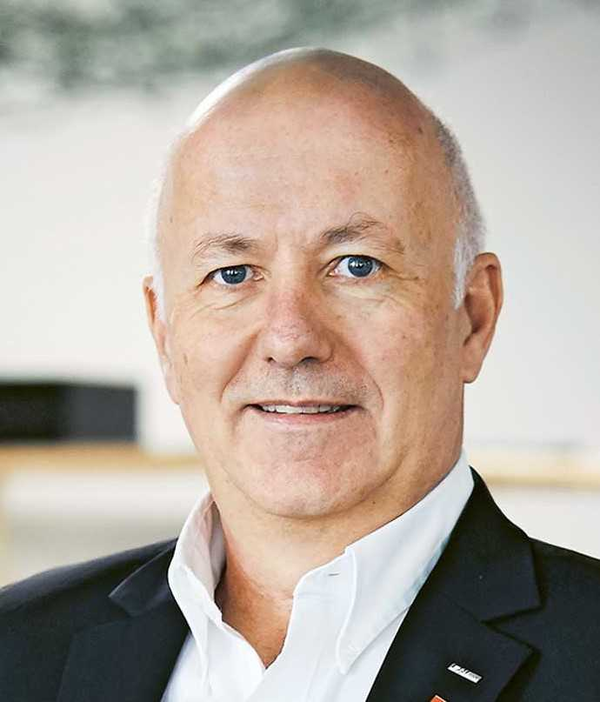}}]{Roland Siegwart} (F'08)
 is professor for autonomous mobile robots at ETH Zurich, founding co-director of the technology transfer center Wyss Zurich and board member of multiple high tech companies. He studied mechanical engineering at ETH Zurich, spent ten years as professor at EPFL Lausanne (1996 – 2006), held visiting positions at Stanford University and NASA Ames and was Vice President of ETH Zurich (2010-2014). He is IEEE Fellow and recipient of the IEEE RAS Pioneer Award and IEEE RAS Inaba Technical Award. He is among the most cited scientist in robots world-wide, co-founder of more than half a dozen spin-off companies and a strong promoter of innovation and entrepreneurship in Switzerland. His interests are in the design, control and navigation of flying, wheeled and walking robots operating in complex and highly dynamical environments.

\end{IEEEbiography}
\vspace{-1.25 cm}
\begin{IEEEbiography}
[{\includegraphics[width=1in,height=1.25in,clip,keepaspectratio]{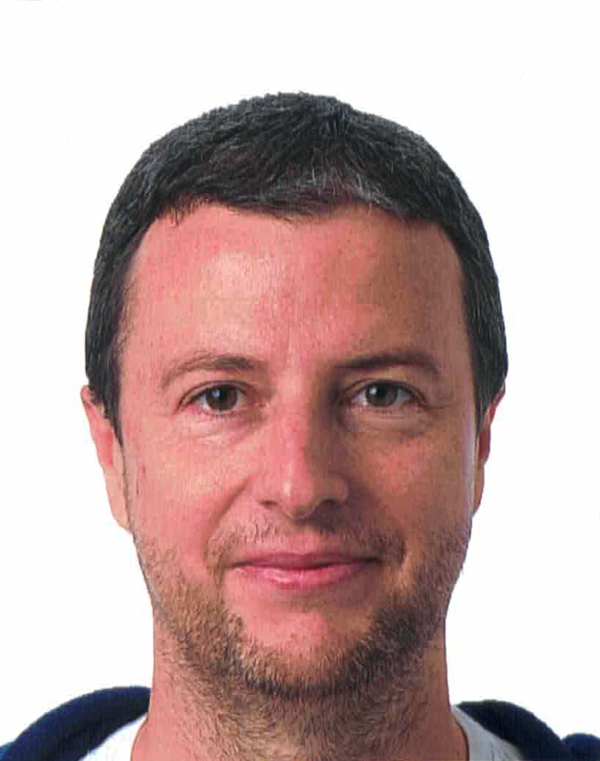}}]{Juan Nieto}
is the Deputy Director at the Autonomous Systems Lab, ETH Zurich. Before joining ETH he was a Senior Research Fellow at the Australian Centre for Field Robotics. He received his Bachelor’s degree in Electronics Engineering from Universidad Nacional del Sur, Argentina, and his PhD in Robotics from the University of Sydney in 2005. His main research interest is in navigation, perception, data fusion and machine learning for mobile robots. He has over 150 scientific publications in international journals and conferences. He has served as Associate Editor for IEEE Robotics and Automation Letters, for IEEE International Conference on Robotics and Automation, IEEE International Conference on Intelligent Robots and Systems, and Robotics Science and Systems, among others. 
\end{IEEEbiography}

\end{document}